\documentclass[10pt,twocolumn,letterpaper]{article}

\usepackage{graphicx}
\usepackage{xcolor} 
\usepackage{amsmath}
\usepackage{amssymb}
\usepackage[T1]{fontenc}
\usepackage{lmodern}
\usepackage{booktabs}
\usepackage{multirow}
\usepackage{graphicx}
\usepackage[margin=0.8in]{geometry}
\usepackage{pifont}
\usepackage{newtxtext,newtxmath}
\usepackage{titlesec}
\usepackage{etoolbox}

\usepackage{pifont}
\usepackage{cuted}
\usepackage{caption}
\usepackage[export]{adjustbox} % already included for valign
\usepackage{subcaption}
\usepackage{mwe}

\makeatletter
\renewcommand{\maketitle}{
  \twocolumn[{%
  \vspace*{-10pt}
  \begin{center}
    {\Large \bfseries \@title \par} % smaller than default Huge
    \vskip 8pt
    {\normalsize
      \@author
    }
    \vspace*{10pt}
  \end{center}
  }]
}
\makeatother

\titleformat{\section}
  {\large\bfseries}{\thesection}{0.6em}{}
\titlespacing*{\section}{0pt}{8pt plus 2pt minus 2pt}{4pt}
\titlespacing*{\subsection}{0pt}{6pt plus 1pt minus 1pt}{3pt}
\titlespacing*{\subsubsection}{0pt}{4pt plus 1pt minus 1pt}{2pt}

\renewcommand{\thesection}{\arabic{section}.}
\renewcommand{\thesubsection}{\thesection\arabic{subsection}.}
\renewcommand{\thesubsubsection}{\thesubsection\arabic{subsubsection}.}

\makeatletter
\renewenvironment{abstract}{
  \centerline{\bfseries Abstract}
  \normalfont\normalsize
  \begin{list}{}{
      \setlength{\leftmargin}{0pt}%
      \setlength{\rightmargin}{0pt}%
  }
  \item[]
}{
  \end{list}
  \vspace{10pt}
}
\makeatother

\usepackage[export]{adjustbox} % already included for valign
\usepackage{graphicx}
\usepackage{subcaption}
\usepackage{mwe}

\definecolor{background}{rgb}{0.2471,0.2706,0.5137}

\usepackage[pagebackref,breaklinks,colorlinks]{hyperref}

\usepackage[accsupp]{axessibility}

\usepackage[capitalize]{cleveref}
\crefname{section}{Sec.}{Secs.}
\Crefname{section}{Section}{Sections}
\Crefname{table}{Table}{Tables}
\crefname{table}{Tab.}{Tabs.}

\begin{document}

\makeatletter
\let\titleold\title
\renewcommand{\title}[1]{\titleold{#1}\gdef\papertitle{#1}}

\def\maketitlesupplementary{%
   \clearpage
   \begingroup
     % Disable section numbering locally
     \setcounter{section}{0}%
     \renewcommand{\thesection}{}%
     \renewcommand{\thesubsection}{}%
     \renewcommand{\thesubsubsection}{}%
     % Typeset the title block
     \twocolumn[{%
       \centering
       \vspace*{1em}%
       {\Large\bfseries \papertitle\par\vspace{0.5em}}%
       {\large Supplementary Material\\[1.0em]}%
     }]%
   \endgroup
}
\makeatother

\title{SSeg: Active Sparse Point-Label Augmentation\\ for Semantic Segmentation}

\author{
Cesar Borja \hspace{1em}
Carlos Plou \hspace{1em}
Ruben Martinez-Cantin \hspace{1em}
Ana C. Murillo\\
DIIS-I3A, University of Zaragoza, Spain\\
{\tt\small \{cborja, c.plou, rmcantin, acm\}@unizar.es}
}
\date{}
\maketitle

%%%%%%%%% ABSTRACT
\begin{abstract}
   Semantic segmentation is essential for automating remote sensing analysis in fields like ecology. However, fine-grained analysis of complex aerial or underwater imagery remains an open challenge, even for state-of-the-art models. Progress is frequently hindered by the high cost of obtaining the dense, expert-annotated labels required for model supervision. 
   While sparse point-labels are easier to obtain, they introduce challenges regarding which points to annotate and how to propagate the sparse information. We present  \textbf{SSeg\footnote{\url{https://sites.google.com/unizar.es/sseg/home}}}, a novel framework that addresses both issues.  SSeg first employs an active sampling strategy to guide annotators, maximizing the value of their point labels. Then, it propagates these sparse labels with a hybrid approach leveraging both the best of SAM2 and superpixel-based methods. Experiments on two diverse monitoring datasets demonstrate SSeg's benefits over state-of-the-art approaches. Our main contribution is a simple but effective interactive annotation tool integrating our algorithms. It enables ecology researchers to leverage foundation models and computer vision to efficiently generate high-quality segmentation masks to process their data.
   %We also release an interactive annotation tool integrating our algorithms, enabling ecology researchers to efficiently generate high-quality segmentation masks.
\end{abstract}

%\vspace{-1em}

%%%%%%%%% BODY TEXT
\section{Introduction}
\label{sec:intro}

Semantic segmentation is key for environmental monitoring, from mapping the seafloor to analyzing aerial images. To work automatically, models typically need detailed pixel-level label supervision. However, fields such as ecology have large datasets of unlabeled images and rarely have dense annotations, as dense pixel‑level annotation is time‑consuming and often requires experts. As a middle point, images are typically annotated with a small number of sparse point‑level annotations, for short, \textit{point-labels}~\cite{raine2022point}. 
For example, the CoralNet project~\cite{beijbom2012automated} has underwater monitoring data labeled by marine biology experts. However, each image is sparsely labeled, with only 50–200 randomly or uniformly sampled labeled pixels per image. This lack of dense annotations hinders the training and subsequent accurate performance of state-of-the-art semantic segmentation models. Different solutions, including the use of superpixels and deep feature affinity models, have emerged to build solutions from point-label annotations. In this context, our work explores two main challenges.

\begin{figure*}[!tb]
    \centering
    % --- First Row with Header ---
    \begin{minipage}[c][0.18\linewidth][c]{0.03\linewidth}
        \centering
        \raisebox{20mm}{\rotatebox{90}{\footnotesize Rand. sampling}}
    \end{minipage}%
    \begin{subfigure}{0.13\linewidth}
        \includegraphics[width=\linewidth]{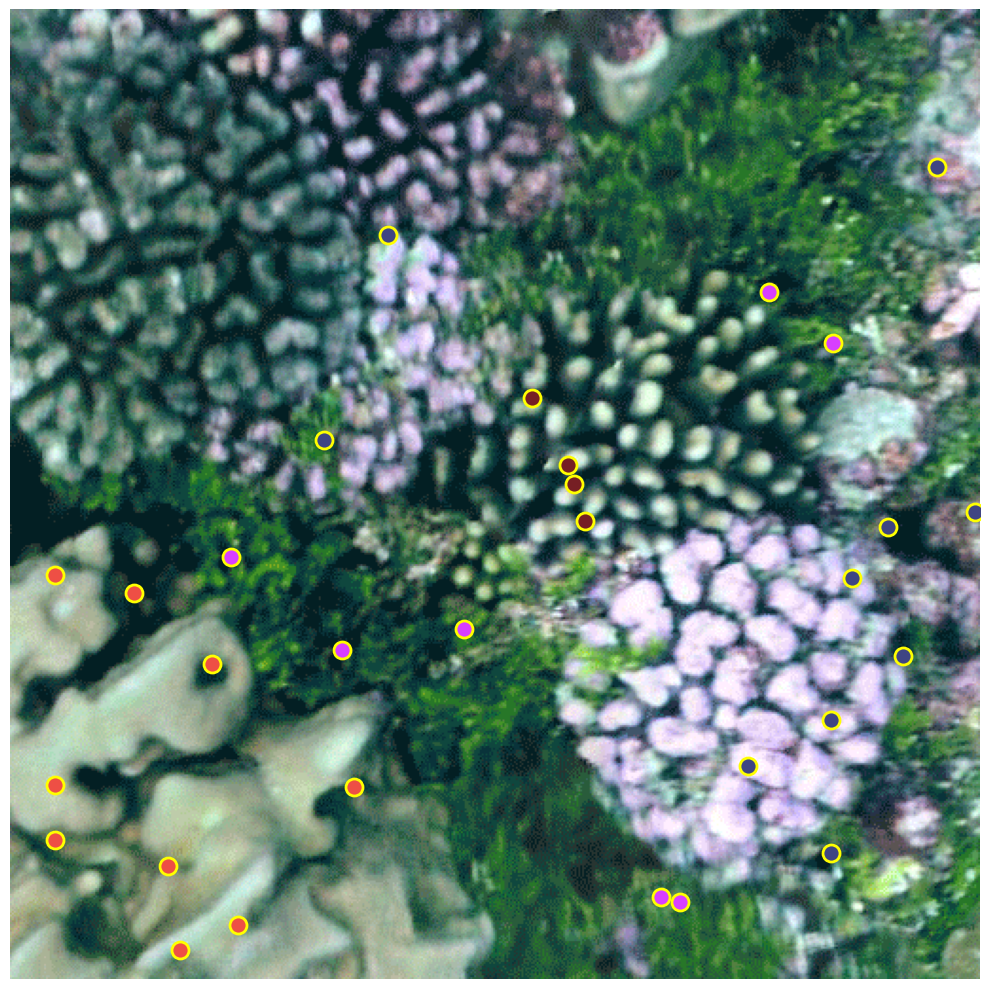}
    \end{subfigure}
    \begin{subfigure}{0.13\linewidth}
        \includegraphics[width=\linewidth]{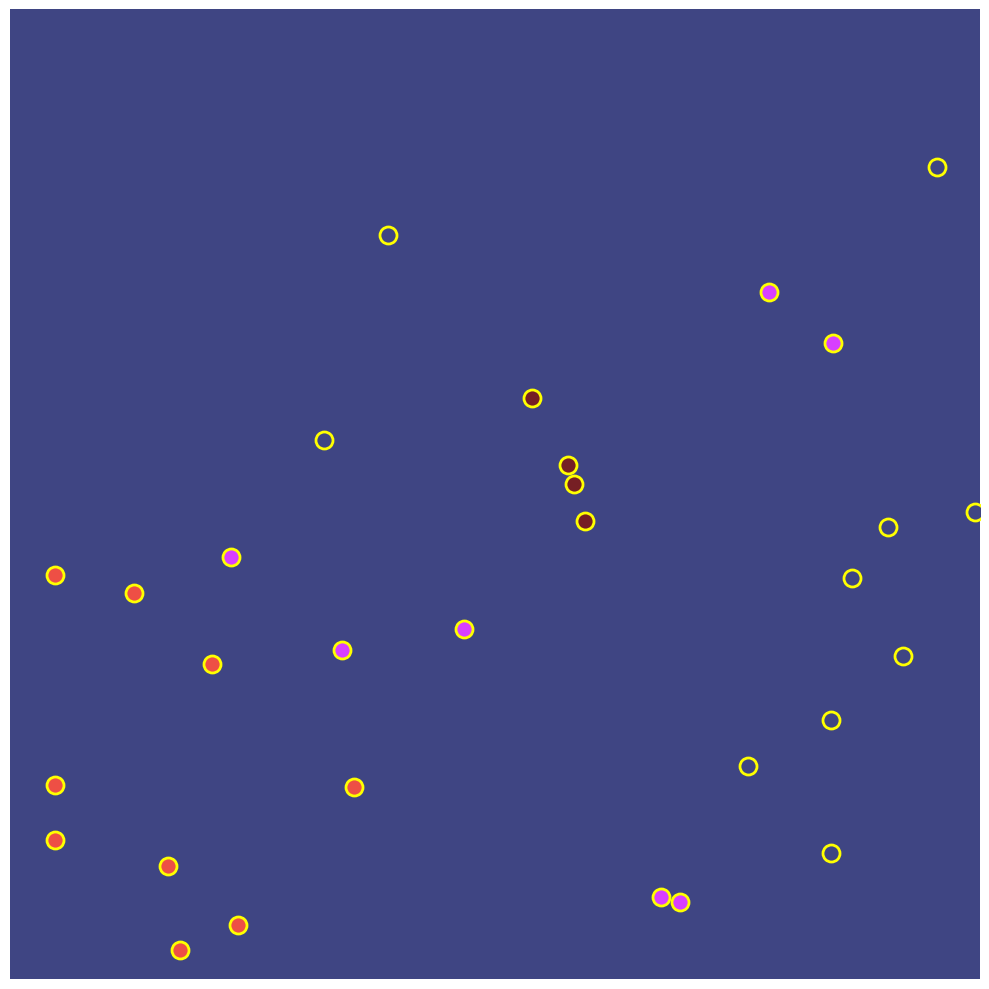}
    \end{subfigure}
    \begin{subfigure}{0.13\linewidth}
        \includegraphics[width=\linewidth]{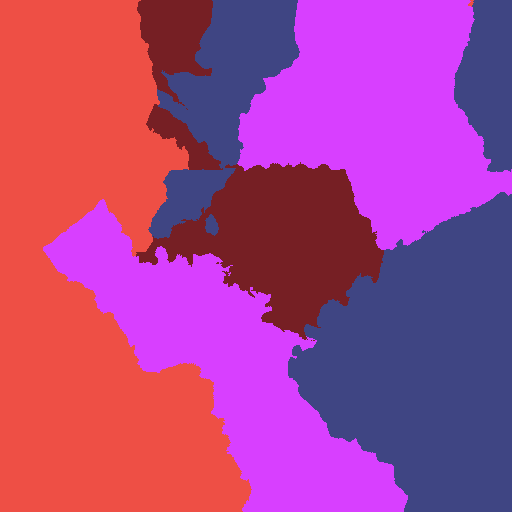}
    \end{subfigure}
    \begin{subfigure}{0.13\linewidth}
        \includegraphics[width=\linewidth]{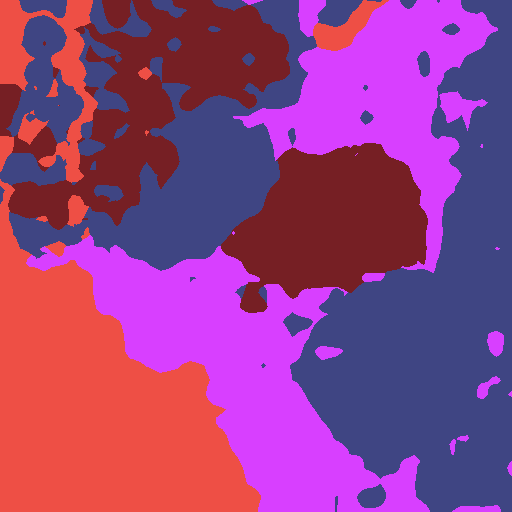}
    \end{subfigure}
    \begin{subfigure}{0.13\linewidth}
        \includegraphics[width=\linewidth]{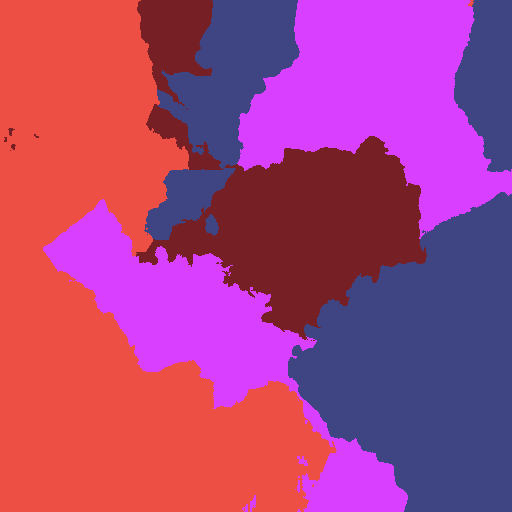}
    \end{subfigure}
    \begin{subfigure}{0.13\linewidth}
        \includegraphics[width=\linewidth]{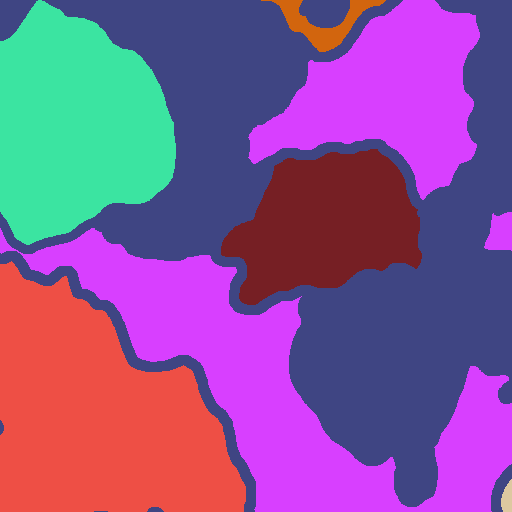}
    \end{subfigure}
\rotatebox{90}{\hspace{.2em}\fcolorbox{black}{background}{\rule{0pt}{6pt}\rule{9pt}{0pt}}\text{\footnotesize{\hspace{0.5em}Background}}}

    \vspace{-14.2mm}

    % --- Second Row with Header ---
    \begin{minipage}[c][0.18\linewidth][c]{0.03\linewidth}
        \centering
        \raisebox{29mm}{\rotatebox{90}{\footnotesize \textbf{Ours sampling}}}
    \end{minipage}%
    \begin{subfigure}{0.13\linewidth}
        \includegraphics[width=\linewidth]{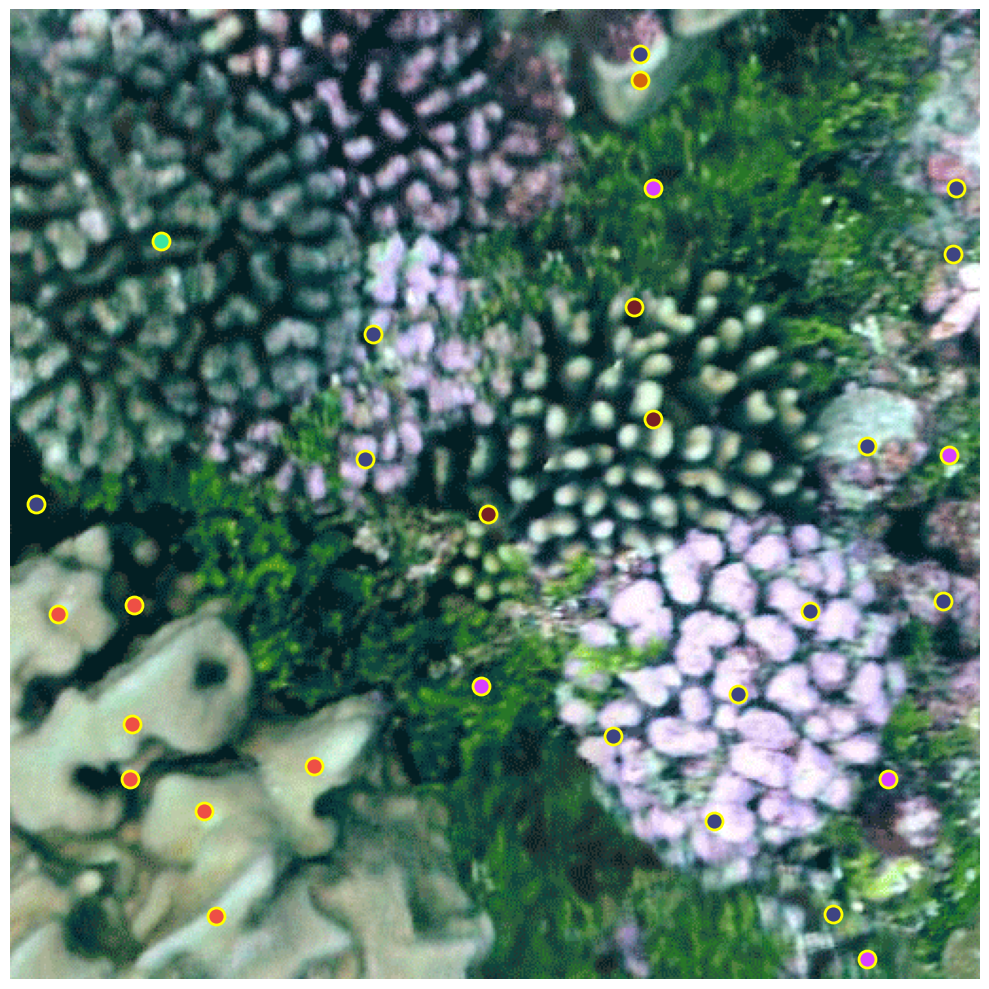}
        \caption*{\footnotesize (a) Image+PL}
    \end{subfigure}
    \begin{subfigure}{0.13\linewidth}
        \includegraphics[width=\linewidth]{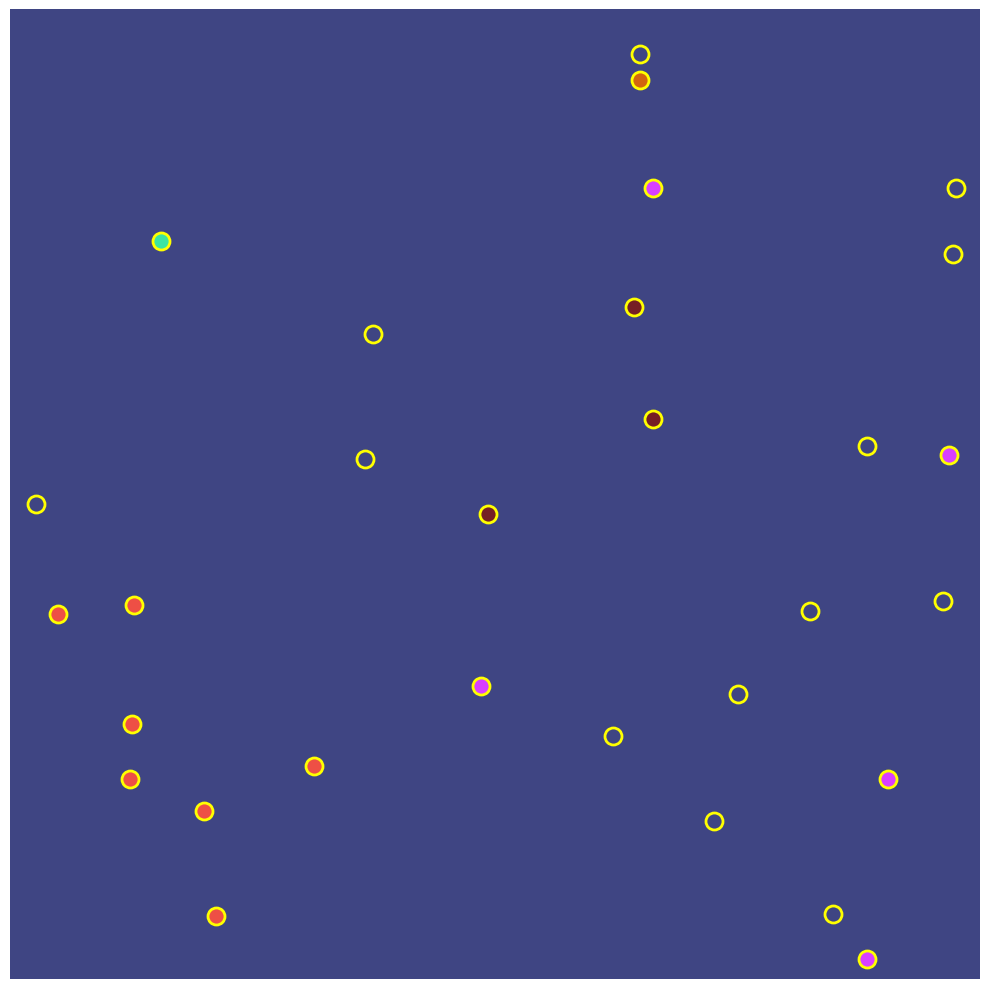}
        \caption*{\footnotesize (b) Point-labels}
    \end{subfigure}
    \begin{subfigure}{0.13\linewidth}
        \includegraphics[width=\linewidth]{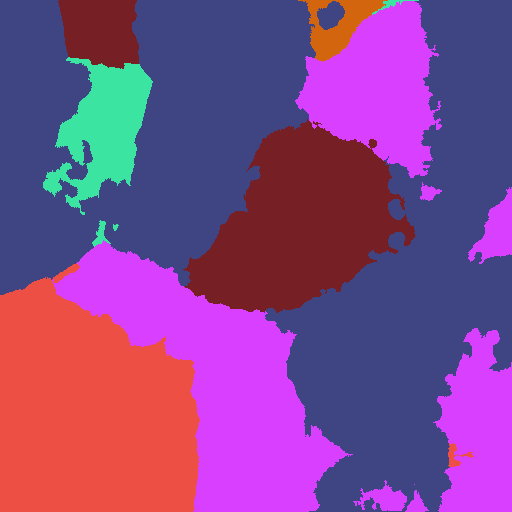}
        \caption*{\footnotesize (c) Superpixel}
    \end{subfigure}
    \begin{subfigure}{0.13\linewidth}
        \includegraphics[width=\linewidth]{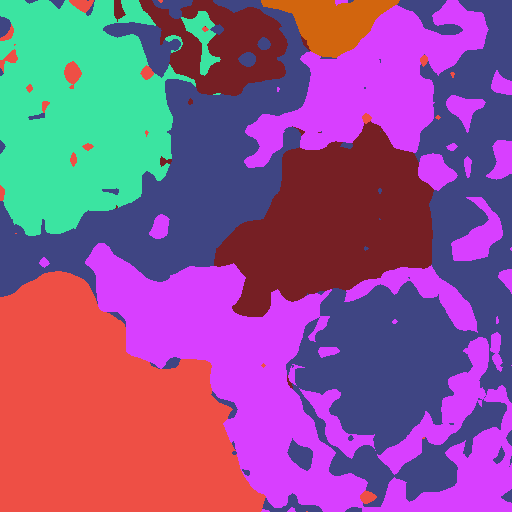}
        \caption*{\footnotesize (d) D+NN}
    \end{subfigure}
    \begin{subfigure}{0.13\linewidth}
        \includegraphics[width=\linewidth]{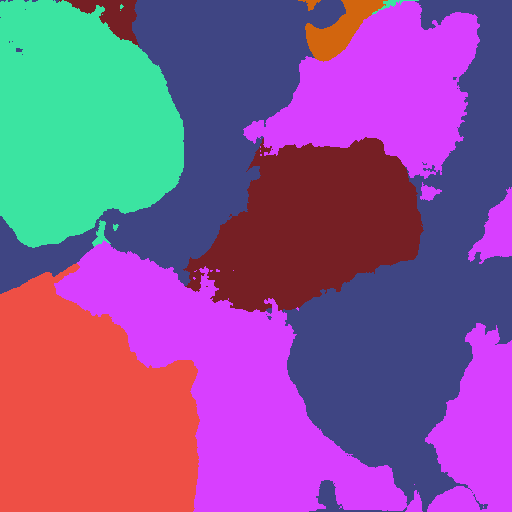}
        \caption*{\footnotesize (e) \textbf{Ours}}
    \end{subfigure}
    \begin{subfigure}{0.13\linewidth}
        \includegraphics[width=\linewidth]{images/teaser/gt.png}
        \caption*{\footnotesize (f) ground truth}
    \end{subfigure}
    \rotatebox{90}{\hspace{.5em}\fcolorbox{black}{background}{\rule{0pt}{6pt}\rule{9pt}{0pt}}\text{\footnotesize{\hspace{0.5em}Background}}}

    \vspace{-14mm}

    \caption{\textbf{Sparse point-label augmentation} of 30 labeled points (yellow circles shown in first two columns) with different approaches. Propagation of labels assigned to \textit{randomly sampled points} (top row) vs. \textit{points sampled by our dynamic strategy} (bottom row). (a) input image with sparse point-labels (PL) superposed, (b) point-labels over background for better visualization, (c) Superpixel-based PLAS~\cite{raine2022point} covers broader regions but often misses fine details and object boundaries, (d) D+NN~\cite{raine2024human} uses global feature matching but suffers from spatial incoherence and texture misclassification. (e) \textbf{SSeg (Ours)} combines SAM2 precision with superpixel coverage to achieve coherent boundaries. Note how our active sampling (bottom row) captures diverse objects missed by random selection, enabling our hybrid augmentation to closely match the (f) Ground Truth.}
    \label{fig:teaser}
\end{figure*}

First, we explore \textit{how to minimize experts annotation efforts.} It is obvious that there is a trade-off between the final dense  segmentation accuracy, and the number and location of the point-labels. But, how many points should the expert label?  Which points are the most interesting ones to know their labels? 
%D+NN~\cite{raine2024human} explores active sampling strategies for interactive point labeling.
Previous works typically assume a uniform point sampling (either random or on a fixed grid) without considering image content. While this simple sampling strategy distributes the points evenly over the image, small or thin objects are often overlooked and some points can land on ambiguous regions (e.g., object boundaries), resulting in incomplete or incorrect propagation. 
%Another straightforward idea would be to place points only on objects suggested by recent foundation models, like SAM2~\cite{ravi2024sam}.  
Our work addresses this concern with our novel active point selection strategy. 

The second challenge to tackle is \textit{how to leverage the point-labels provided by the expert.} A natural solution explored in previous work to leverage sparse point-labels is label augmentation, where the sparse point‑labels are propagated into full‑image masks with promising results~\cite{raine2022point, raine2024human}. The resulting dense semantic segmentations not only serve for valuable expert analysis on their own, but also as training data for conventional segmentation models. 
Most common strategies for label augmentation are based on superpixel image segmentation, or on more recent vision foundation models such as DINO2~\cite{oquab2023dinov2} or SAM2.
% Superpixel segmentation is widely used and guarantees every pixel receives a label but often produces coarse region boundaries that miss fine object details. 

% SAM2 is able to produce precise masks from individual point queries but do not ensure full coverage with only a few seed points. DINOv2 features are used in D+NN~\cite{raine2024human} to achieve the overall state-of-the-art for sparse point-label augmentation on the challenging coral reef UCSD Mosaics~\cite{edwards2017large} dataset, by computing per‑pixel affinities to guide the label propagation. 
However, neither approach offers a complete solution: superpixels often miss fine object details, while foundation models can struggle with dense coverage and spatial coherence. 
SSeg is an alternative to existing approaches that efficiently leverages the advantages of both strategies to achieve the best performance.

To sum up, we present \textbf{SSeg}, a practical workflow that builds upon existing point-label augmentation methods to reduce annotation efforts by ecology experts, with two key components (illustrated in Fig.~\ref{fig:teaser}): 1) a \textbf{hybrid sparse label augmentation strategy} that combines superpixel- and SAM2-based segmentations; 2) an \textbf{active point sampling} strategy to better guide expert annotations; and 3) a validation of our pipeline for\textbf{ downstream model training}. Our experiments demonstrate +3--8\% mIoU gains over baselines, and near-full recovery of downstream performance with four orders of magnitude fewer point-labels.

To support adoption by ecology experts, we release a simple \textbf{interactive annotation tool} based on SSeg to generate dense pixel-level labels for their own data.

\section{Related work}
\label{sec:related}
\paragraph{Semantic Segmentation in natural environments.}
Semantic segmentation assigns a class label to every pixel in an image, enabling fine‑grained scene understanding. Early deep models such as Fully Convolutional Networks (FCNs)\cite{long2015fully} and U‑Net\cite{ronneberger2015u} delivered end‑to‑end pixel‑wise predictions, while the DeepLab family~\cite{chen2017deeplab} introduced multi‑scale context modules to better capture object boundaries. More recent transformer‑based architectures~\cite{xie2021segformer, liu2021swin} improve global context modeling, and vision foundation models like DINOv2~\cite{oquab2023dinov2}, SAM~\cite{kirillov2023segment}, and SAM2~\cite{ravi2024sam} provide promptable or self‑supervised mask proposals with minimal task‑specific training.

Natural scenarios often present more challenging use cases. For example, underwater scenarios come with additional challenges~\cite{mobley1994light, berman2020underwater, bekerman2020unveiling} (color distortion, scale variations, and complex textures) which make standard segmentation models struggle. To fix this, researchers have adapted many architectures for marine scenes~\cite{he2024uiss, chicchon2023semantic, wei2022image, liu2020semantic, zhang2022dpanet}, incorporating domain‑specific enhancements. More recently, CoralSCOP~\cite{Zheng_2024_CVPR} builds on SAM to add a semantic head for coral masks in a zero‑shot setting. These models are typically trained on specialized datasets like SUIM~\cite{islam2020semantic} and USOD10K~\cite{hong2023usod10k}, which include diverse images from general underwater expeditions, or SkyScapes~\cite{azimi2019skyscapes}, which provides high-resolution annotations for aerial lane detection. In contrast, UCSD Mosaics~\cite{edwards2017large} provides dense, multi-species coral imagery for semantic segmentation. Despite this progress, dense pixel‑level labels remain scarce, especially in benthic environments, where annotation is costly and done by marine biology experts, as in the Eilat dataset~\cite{beijbom2016improving}, which includes coral images annotated with  sparse point-labels. Similarly, CoralNet~\cite{beijbom2012automated} provides a vast collection of monitoring images from many marine sites, but each image contains only a small set of sparse point-labels. We leverage such data by sparse annotations expansion into full-image segmentations through label augmentation.

\paragraph{Label Augmentation.}
Superpixel-based methods remain a successful alternative for ecological monitoring in non-standard domains. CoralSeg~\cite{alonso2019coralseg, alonso2017coral} applies a multi-level strategy using SEEDs~\cite{van2015seeds}, later improved in \cite{pierce2020reducing} by replacing SEEDs with optimized SLIC~\cite{achanta2012slic} superpixels for better consistency. The Superpixel Sampling Network (SSN)~\cite{jampani2018superpixel} introduced differentiable, trainable superpixels, inspiring other CNN-based methods~\cite{wang2020end, cai2021revisiting, tu2018learning} adapted for point-label propagation. More recently, PLAS~\cite{raine2022point} introduced a loss-based optimization aligning boundaries with visual features while minimizing class conflicts. In the aerial domain, methods like FESTA~\cite{hua2021semantic} [cite: 3, 9] have demonstrated that sparse scribbles can be effectively utilized by regularizing feature and spatial relations. Beyond superpixels, Zhang et al.~\cite{zhang2024point} combine SegFormer~\cite{xie2021segformer} with SAM-based augmentations for seagrass mapping, while D+NN~\cite{raine2024human} uses DINOv2~\cite{yang2024denoising} feature affinity. Our work leverages both foundation models and superpixels to build a more robust strategy. Nevertheless, effective augmentation relies significantly on point placement, making active selection essential.

\paragraph{Active Point Selection.}
Active Learning (AL) is an iterative strategy in which a model selects the most informative unlabeled samples for annotation to improve performance under a limited labeling budget~\cite{settles2009active, lopes2014active, mosqueira2023human, fu2013survey, olsson2009literature}. When this selection relies on deep neural networks and their uncertainty or feature representations, it is known as Deep Active Learning (DAL)~\cite{ren2021survey, zhan2022comparative, li2024survey, siddiqui2020viewal}. DAL for semantic segmentation aims to reduce labeling cost by selecting the most informative pixels or regions to annotate. Many DAL approaches operate at the image level, ranking entire images by uncertainty or diversity to choose which ones to. Xie et al.~\cite{Xie_2022_CVPR} observe that labeling whole images is wasteful, and propose a region-based strategy that selects connected regions of high uncertainty and class impurity. Another baseline based in point-based selection is LabOR~\cite{shin2021labor}, which queries a few pixels the model is uncertain about, but this can be inefficient in complex scenes. Kim et al.~\cite{kim2023adaptive} go further by querying adaptive superpixels: at each round they merge pixels of similar features into segments and request one label per superpixel. In medical imaging,~\cite{qu2023abdomenatlas} applied active learning to multi-organ CT annotation: by combining pre-trained models with attention-guided corrections, and attention-guided corrections, they labeled 8,000 volumes in 3 weeks with minimal mistakes.

% This superpixel-based AL drastically cuts annotator effort (one click per segment) while extracting noisy labels. In medical imaging,~\cite{qu2023abdomenatlas} applied active learning to multi-organ CT annotation: by combining multiple pre-trained models, error detection, and attention-guided corrections, they labeled 8,000 volumes in 3 weeks with minimal mistakes.

Recent remote sensing approaches, such as HSLabeling~\cite{lin2025hslabeling}, go a step further by adaptively selecting hybrid sparse labels (e.g., points vs. blocks) to minimize cost. However, we restrict our focus to optimizing point-labels to align with rapid click-based annotation workflows standard in marine surveys. Although related, most AL scenarios focus on minimizing model uncertainty by sampling ambiguous points. Instead, our goal is to maximize the information in the expanded masks, requiring high-quality seeds. D+NN~\cite{raine2024human} attempts this via DINOv2 similarity but relies on expert-provided seeds. In contrast, our method automatically proposes all points, allowing experts to focus solely on assigning class labels.

\section{SSeg}
SSeg is a point-label augmentation method that produces complete image segmentations from just a few sparse point-labels that serve as seeds. 
 
Our method includes a novel active point selector that strategically selects the most informative points (pixels) to query for their labels by balancing likely  object regions and background areas, improving segmentation quality while minimizing annotation effort.

\textbf{Problem Definition.}  
Given an image \( I \), a set of semantic labels \( L \), and a labeling budget of \( n \) pixels from which we can query their ground-truth labels, our goal is to produce a full-image semantic segmentation \( I_L \). Formally, we construct a set of point–label pairs%
\begin{equation}
P_L = \{(p_1, l_1), (p_2, l_2), \dots, (p_n, l_n)\},
\end{equation}%
where each \(p_i \in I\) is a pixel coordinate and \(l_j\in L\) its semantic label, and we aim to infer a dense labeling function $S : I \;\to\; L$, assigning every pixel in \(I\) to one of the $m$ classes in \(L\), and satisfying
\begin{equation}
S(p_i) \;=\; l_j,
\quad\forall\;(p_i, l_j)\in P_L.
\end{equation}
We will refer to the full-image semantic segmentation as $I_L=S(I)$.

\subsection{Method Overview}
SSeg produces a full-image semantic segmentation $I_L$ by combining two main components: (i) an active point selection strategy that chooses which are the $n$ most informative pixels $p_i$ to query for their labels $l_j$, and (ii) a point label augmentation method that propagates each point-label $(p_i, l_j) \in P_L$ to a dense labeled mask $M_i$ with label $l_j$.

\paragraph{Active Point Selection.} 
The quality of the final semantic segmentation depends on which point-labels $P_L$ are selected and used as seeds for the augmentation stage. To maximize their impact, we design a hybrid sampling strategy that combines a novel active method guided by an acquisition function with random sampling (Sec. \ref{sec:active_step}). The acquisition function scores pixels based on proximity to object centroids (from SAM2 \cite{ravi2024sam} masks) and distance to previously labeled points, promoting both object-relevant supervision and spatial coverage. The remaining points are selected from regions not covered by any detected object mask to ensure background supervision, resulting in a balanced and informative label set. We perform this phase last to avoid penalizing masks surrounded by background.

\paragraph{Point Label Augmentation.} To obtain the full-image semantic segmentation $I_L$, we combine two complementary strategies that propagate the point-labels $P_L$ into dense labeled masks (Sec. \ref{sec:augmentation_step}). First, each point is expanded into a mask using SAM2 \cite{ravi2024sam}, producing high-quality object boundaries. These masks are carefully merged to construct a partial segmentation. In parallel, PLAS~\cite{raine2022point}—a superpixel-based label propagation method—is applied to infer a dense prediction, filling in the regions that SAM2 leaves unlabeled to ensure full image coverage.

\subsection{Active Point Sampling}
\label{sec:active_step}

Our first goal is to select the most informative set of point--label pairs \(P_L\), which will serve as seeds for the subsequent label propagation stage. We define an active sampling strategy guided by three criteria:  
(1) \textit{Proximity to object centroids}, to avoid placing point-labels on object boundaries;  
(2) \textit{Coverage}, to encourage spatial distribution of labels across the image;  
(3) \textit{Background inclusion}, to query labels in low-contrast or background regions where segmentation models, which are typically optimized for object-centric masks, tend to fail.

To fulfill these criteria, we divide the budget of \(n\) point-labels between two strategies. The first half is selected sequentially via an acquisition function \(A(p)\), which scores each pixel \(p \in I\) based on its proximity to object centroids \(O(p)\) and its spatial distance to already selected points \(E(p)\):

\begin{equation}
A(p) = \lambda\,O(p) + (1 - \lambda)\,E(p),
\end{equation}

where \(\lambda \in [0, 1]\) controls the trade-off between both terms. This acquisition function \(A(p)\) is updated dynamically each time a new label \((p_i, l_j)\) is added to \(P_L\).

\paragraph{Proximity to object centroids.}  
We run SAM2~\cite{ravi2024sam}, which uses an internal random grid of points to automatically generate candidate object masks. For each mask, we compute its centroid \(c_k\) and area \(a_k\). The proximity term \(O(p)\) assigns higher scores to pixels located near the centroids of larger masks:

\begin{equation}
O(p) = \frac{\sum_k a_k \cdot \left(1 - \frac{\|p - c_k\|}{d_{\max}} \right)}{\sum_k a_k},
\end{equation}

where \(d_{\max} = \sqrt{H^2 + W^2}\) is the image diagonal, used for distance normalization. Note that \(O(p)\) is fixed and computed only once.

\paragraph{Exploration coverage.}  
This term promotes the selection of pixels far from already queried points \(p_i \in P_L\). If no labels have been queried yet, we define \(E(p) = 0\). Otherwise,

\begin{equation}
E(p) = \frac{1}{d_{\max}} \min_{p_i \in P_L} \|p - p_i\|.
\end{equation}

\paragraph{Background sampling.}  
The remaining half (see Table \ref{tab:ablations_combined} for the ablation of the random sampling ratio) of the point-label budget is randomly selected from areas not covered by any SAM2~\cite{ravi2024sam} mask (as determined during centroid proximity computation). These regions typically correspond to background or low-contrast zones where segmentation models underperform. Querying labels in such regions can significantly improve the downstream label propagation quality.

\subsection{Point Label Augmentation}
\label{sec:augmentation_step}
To produce the full-image semantic segmentation $I_L$ from the sparse point-labels $P_L$, we follow a two-step augmentation strategy.

\paragraph{SAM2 Expansion and Overlap Solving.}
Each labeled point $p_i$ is expanded into a binary mask $M_i \subseteq I$ using SAM2~\cite{ravi2024sam}, assigning its corresponding label $l_j$ to all pixels in $M_i$. However, as masks may overlap, some pixels may belong to multiple masks and thus have conflicting labels. We resolve these conflicts by computing a unique label per pixel. 
For each pixel $p \in I$, we define the set of mask indices $S_p$ covering it:
\begin{equation}
S_p = \{i \mid p \in M_i \} \subseteq \{1,\hdots,n\}.
\end{equation}%
\noindent If $|S_p| = 1$, we directly assign the label $l_i$ to $p$. If $|S_p| > 1$, we compute the overlap region $O_p= \bigcap_{i \in S_p} M_i$ for those masks.

We then evaluate which label is most supported in $O_p$ based on proximity to point-labels. With $c_p$ the centroid of $O_p$, we get a score for each mask $i \in S_p$:
\begin{equation}
  \mathrm{Score}(p, i)
  = \sum_{\substack{(p_j, l_j)\in P_L \\ p_j\in O_p \\ l_j=l_i}}
    \frac{1}{1 + \|p_j - c_p\|}.
\end{equation}
\noindent This score quantifies the support that label $l_i$ has within $O_p$, giving more weight to nearby points. Each pixel in $O_p$ is then assigned the label \(l_{i^*}\) of the mask with the highest score, where \(i^* = \arg\max_{i \in S_p} \mathrm{Score}(p, i)\). This process yields a partial segmentation map $I^{SAM2}_L$, labeling all pixels covered by at least one SAM2 mask. Pixels with $S_p = \emptyset$ remain unlabeled.
\vspace{-0.3cm}
\paragraph{PLAS Filling.}
In parallel, we run PLAS~\cite{raine2022point} on the original image using the same point-labels \(P_L\), producing a superpixel-based segmentation map \(I^{PLAS}_L\) that labels the entire image. For any pixel \(p\) not labeled in $I^{SAM2}_L$, we use the label from \(I^{PLAS}_L\). The final segmentation map is:%
\begin{equation}\label{eq:final_seg}
  I_L(p) = 
  \begin{cases}
    I^{SAM2}_L(p), & S_p \neq \emptyset,\\ %
    I^{PLAS}_L(p), & S_p = \emptyset, %
  \end{cases}
\end{equation}%
\noindent ensuring full label coverage of the image.

\section{Experiments}

\subsection{Experimental setup}
All experiments were conducted using an NVIDIA GeForce RTX 4090 GPU. Unless otherwise stated, all inference times refer to execution on this hardware.

\subsubsection{Datasets used}
%\vspace{-0.1cm}
\paragraph{UCSD Mosaics~\cite{edwards2017large}} consists of 16 large mosaics of multi-species coral reefs, each with a resolution exceeding 10K $\times$ 10K. CoralSeg\cite{alonso2019coralseg} introduced a cropped version of this dataset by dividing the mosaics into 512 $\times$ 512 resolution images, which we adopt in our work. The dataset has 35 semantic classes, including background class, and provides dense label masks.
\vspace{-0.3cm}
\paragraph{SkyScapes~\cite{azimi2019skyscapes}} consists of 12 large-scale aerial mosaics (5616 $\times$ 3744). We utilize the 10 training and validation images, containing 21 semantic classes, as the test set lacks public ground truth. We divide each mosaic into a 6 $\times$ 6 grid to produce 36 patches per image (936 $\times$ 624). We chose this resolution to preserve sufficient spatial context for structural understanding.

\subsubsection{Baselines and configuration}
%\vspace{-0.1cm}
\paragraph{PLAS - Ens~\cite{raine2022point}.} We use the default PLAS configuration for comparison: Same Gaussian normalization and conflict loss terms, distortion loss computed with 3000 pixels, 100 superpixels per image and SSN~\cite{jampani2018superpixel} pretrained on the UCSD Mosaics dataset. Also, we use the `ensamble' mode that merges three different outputs. 
\vspace{-0.9cm}
\paragraph{D+NN~\cite{raine2024human}.} D+NN is our main baseline, given its similarity to our approach and its state-of-the-art performance in sparse point-label augmentation.
\vspace{-0.5cm}
\paragraph{CoralSCOP~\cite{Zheng_2024_CVPR}.} CoralSCOP is a SAM-based model fine-tuned for coral segmentation, so we evaluate it exclusively on the UCSD Mosaics dataset. We utilize the official implementation and pre-trained weights using the default configuration provided by the authors.
\vspace{-0.5cm}
\paragraph{SSeg-SAM2Only (Ours).} This refers to our approach explained in section \ref{sec:augmentation_step} but without including the PLAS filling. We employ the large SAM2.1 checkpoint and configure the $\mathrm{SAM2AutomaticMaskGenerator}$ with the following parameters to produce large and good quality masks: $\mathrm{points\_per\_side}$ = $28$, $\mathrm{points\_per\_batch}$ = $512$, $\mathrm{pred\_iou\_thresh}$ = $0.5$, $\mathrm{stability\_score\_thresh}$ = $0.9$, $\mathrm{stability\_score\_offset}$ = $0.7$, $\mathrm{mask\_threshold}$ = $0.25$, $\mathrm{box\_nms\_thresh}$ = $0.45$, $\mathrm{crop\_n\_layers}$ = $0$, $\mathrm{min\_mask\_region\_area}$ = $1500$, and $\mathrm{multimask\_output}$ = True.
\vspace{-0.5cm}
\paragraph{SSeg (Ours).} This refers to the full version of our approach explained in section \ref{sec:augmentation_step}, including PLAS filling. We use the same PLAS and SAM2 configurations as previously described.

We also fix $\lambda = 0.5$ for all experiments involving our active point-sampling approach to give equal importance to exploration and object proximity, an ablation of $\lambda$ is available in Table~\ref{tab:ablations_combined}.

\subsubsection{Evaluation}
One of the key goals in ecological studies is to monitor the coverage of certain species~\cite{mcclenachan2017ghost} due to habitat loss. We measure the mean Pixel Accuracy (\textbf{mPA}) and mean Intersection over Union (\textbf{mIoU}), two commonly used metrics~\cite{alonso2019coralseg, raine2022point, raine2024human} that reflect how well the segmentation captures species presence in the image. Let $\text{TP}_l$, $\text{FP}_l$, and $\text{FN}_l$ represent the true positive, false positive, and false negative pixels for a given semantic label $l\in L$. 
The per-class pixel accuracy is: $\text{PA}_l = \frac{\text{TP}_l}{\text{TP}_l + \text{FN}_l}$. The mPA is the average of the $\text{PA}_l$ for all $L$. The mIoU provides a more strict evaluation of segmentation by measuring the overlap between predicted and ground truth regions. The IoU for a single semantic label $l$ is given by $\text{IoU}_l = \frac{\text{TP}_l}{\text{TP}_l + \text{FP}_l + \text{FN}_l}$. 
It computes the ratio between the area of intersection and the area of union. The mIoU is the average of the $\text{IoU}_l$ for all $L$. 

\begin{figure}[!tb]
\rotatebox{90}{\hspace{2em}\fcolorbox{black}{background}{\rule{0pt}{6pt}\rule{9pt}{0pt}}\text{\footnotesize{\hspace{0.5em}Background}}}
    \begin{subfigure}{0.3\linewidth}
        \includegraphics[width=\linewidth]{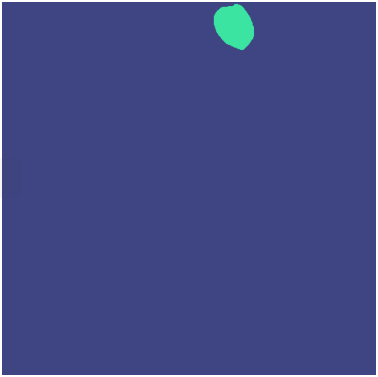}
        \caption{ground truth}
    \end{subfigure}
    \begin{subfigure}{0.3\linewidth}
        \includegraphics[width=\linewidth]{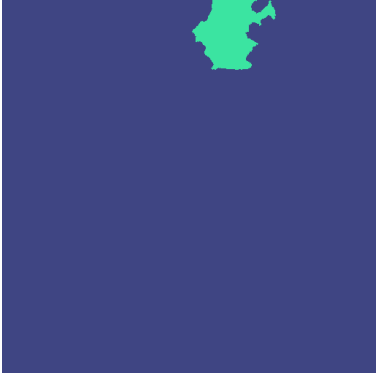}
        \caption{IoU = $35\%$}
    \end{subfigure}
    \begin{subfigure}{0.3\linewidth}
        \includegraphics[width=\linewidth]{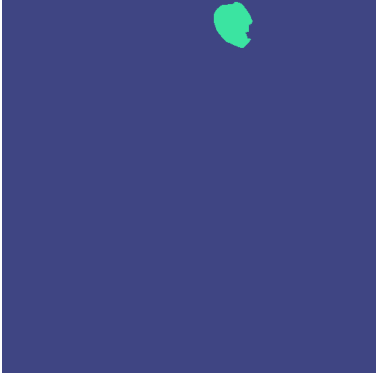}
        \caption{mask-IoU=$93\%$}
    \end{subfigure}

    \caption{Intersection over Union (IoU) vs masked-IoU. (a) ground truth with a small object surrounded by background. (b) predicted mask that oversegments into background, resulting in a low IoU (35.60\%). (c) visual representation of the same predicted mask when background regions are masked out, illustrating what masked-IoU effectively measures. This example illustrates how masked metrics focus on foreground accuracy without penalizing background oversegmentation.}
    \label{fig:standard_vs_masked}
\end{figure}

In practice, we first calculate the number of correctly and incorrectly predicted pixels per class across the entire dataset and then average the resulting scores to obtain the overall mPA and mIoU. We also include ``background" class in all our measurements. 
In addition to the standard metrics, we also report \textbf{masked-mPA} and \textbf{masked-mIoU}, which are computed in the same way but exclude all background-class pixels from the evaluation. This highlights how well the model performs on foreground classes, while standard mPA and mIoU provide a more balanced evaluation by including background pixels as well (see Fig.~\ref{fig:standard_vs_masked}).

\subsection{Results}
The following experiments evaluate the performance of the presented approach for dense annotation of images. Figure~\ref{fig:sparse_sampling_multi} presents some qualitative results, and the supplementary material includes additional ones with the second benchmark. The following experiments illustrate the benefits of our complete approach with respect to existing methods, and then we analyze in more detail the two main components: the point-label augmentation strategy and the active point selection. Finally, we assess the quality of our augmented labels for training a downstream segmentation model.

%\subsubsection{Label augmentation evaluation.}
%First we evaluate the point-label augmentation module, given a set of sparse point-labels for each image, in two different experiments. 

\begin{figure}[!tb]
    \centering
    \includegraphics[width=\linewidth]{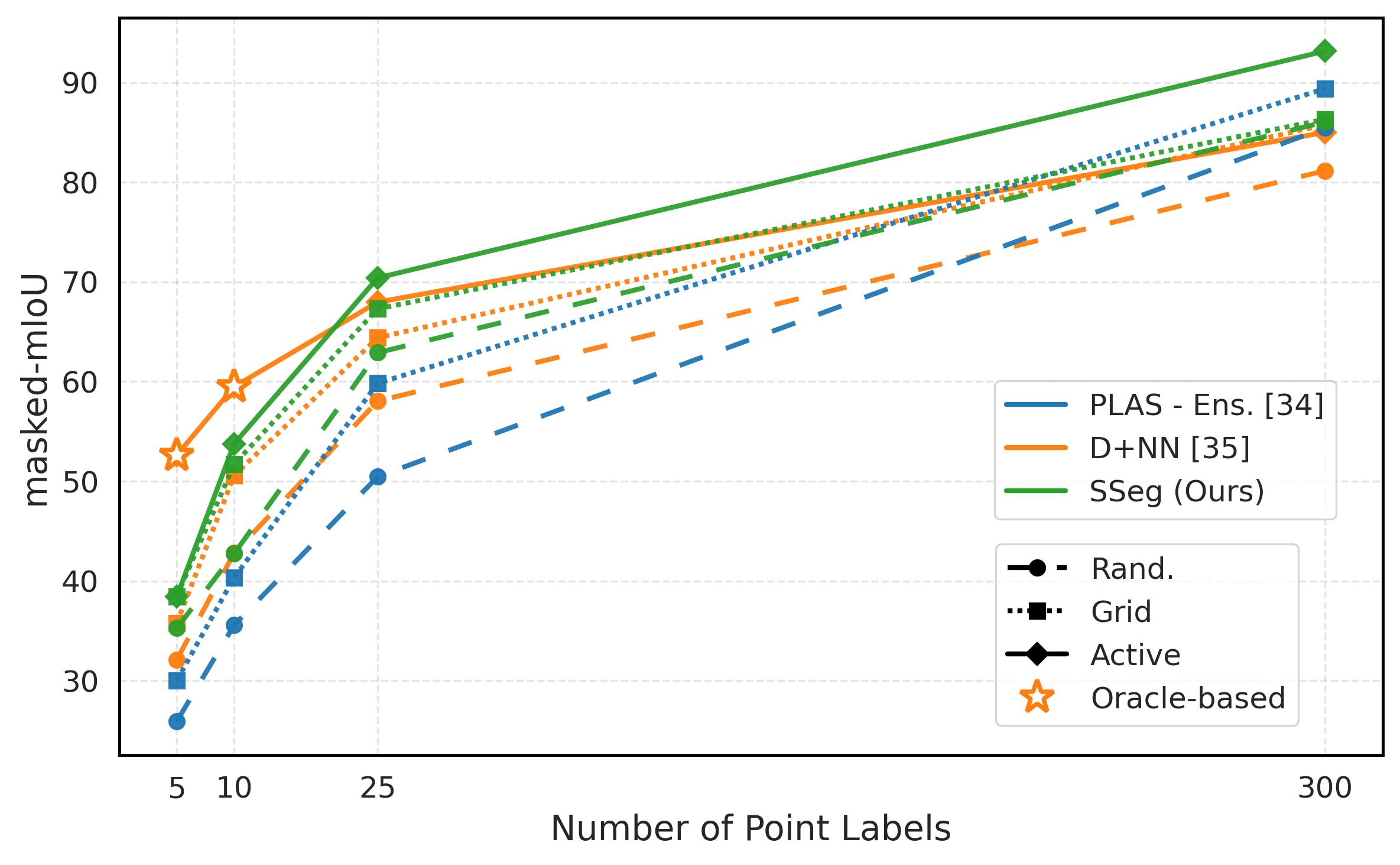}
    \caption{\textbf{Performance analysis on underwater scenes (UCSD Mosaics).} We benchmark our SSeg method against reported results for PLAS and D+NN from~\cite{raine2024human}, using random, grid, and active sampling strategies. The plot shows masked-mIoU (excluding background) relative to the number of point labels. Note that D+NN uses oracle-based sampling (indicated by the star marker) to place the initial 10 points using GT guidance in all configurations. While this privilege creates a noticeable performance gap at 5 and 10 points, SSeg (green) surpasses D+NN at 25 and 300 points without relying on any GT guidance for point placement.}
    \label{fig:results_relatedwork}
\end{figure}

%\vspace{-0.3cm}
%\paragraph{Experiment 1: state-of-the-art comparison.}
\subsubsection{State-of-the-art comparison}
We compare our method with state-of-the-art approaches on sparse label augmentation, PLAS~\cite{raine2022point} and D+NN~\cite{raine2024human}, on the UCSD Mosaics underwater dataset. We follow the exact same settings from the original papers to ensure a fair comparison. Figure~\ref{fig:results_relatedwork} shows the masked-mIoU results for random, grid, and active sampling using 5, 10, 25, and 300 points, as reported in the original works, compared to SSeg results.

A key variation for the sparse label propagation is the how the sparse set of points is selected. We evaluate both static sets of points and active strategies to iteratively find the most interesting point labels to expand. 

For static sampling strategies (random and grid), our method scores highest at 5, 10, and 25 points, averaging +2.38\% masked-mIoU over D+NN. For active sampling strategies, D+NN leads when sampling only 5 or 10 points; however, we should note that this approach relies on what we refer to as `Oracle-based' initialization, where the first 10 points are placed at the centroids of the largest ground-truth objects. SSeg operates without this ground-truth guidance. Despite this, our method surpasses D+NN as the budget increases (+3\% at 25 points, +8\% at 300 points), demonstrating a more effective active strategy to select points to annotate and expand that does not require manual object selection.

\begin{figure}[!tb]
    \centering
    % Row 1
    \begin{subfigure}{0.18\linewidth}
        \includegraphics[height=1.35cm]{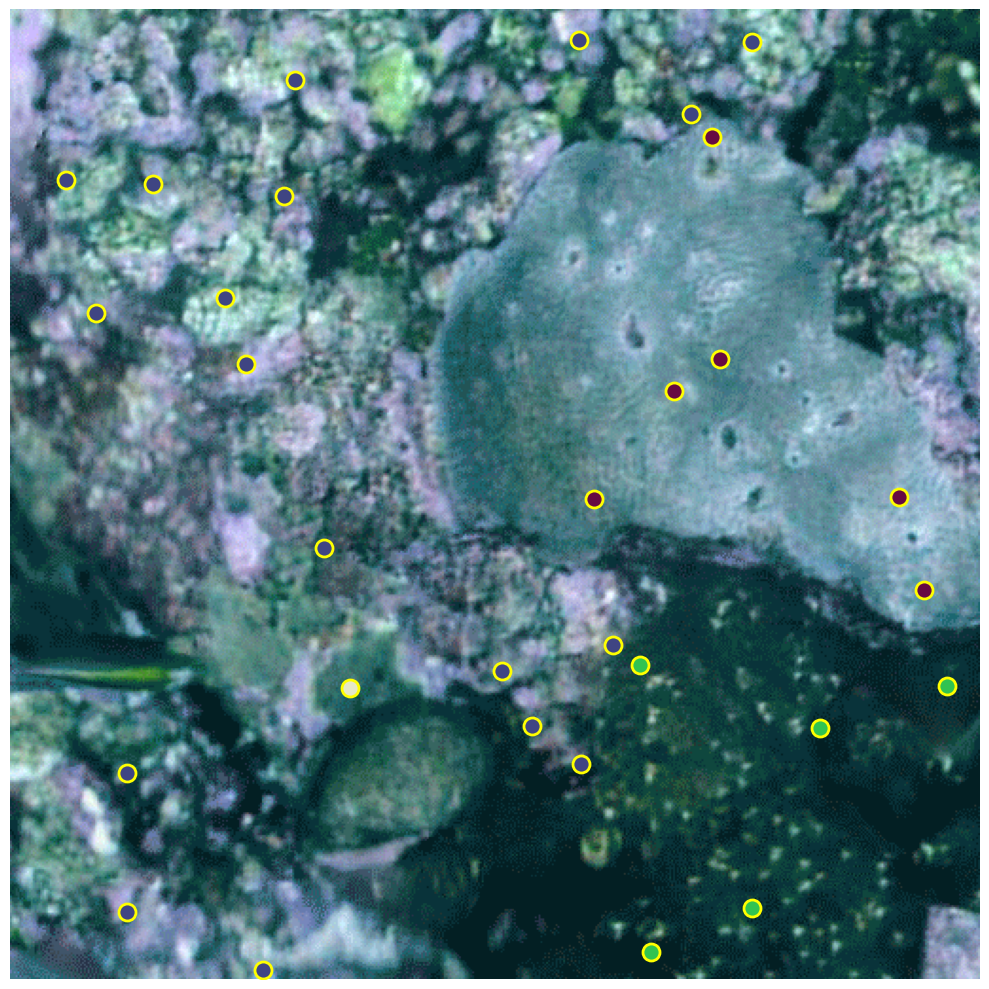}
    \end{subfigure}
    \begin{subfigure}{0.18\linewidth}
        \includegraphics[height=1.35cm]{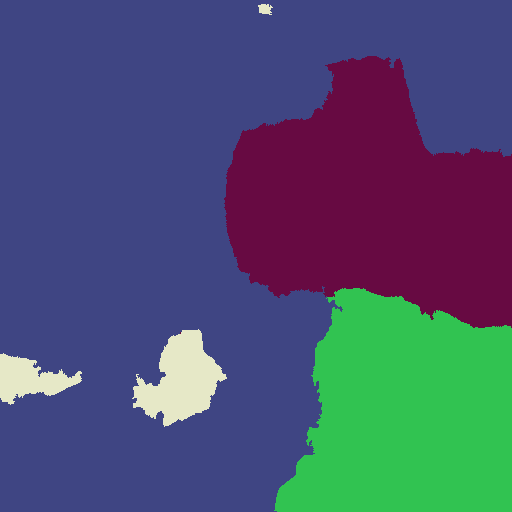}
    \end{subfigure}
    \begin{subfigure}{0.18\linewidth}
        \includegraphics[height=1.35cm]{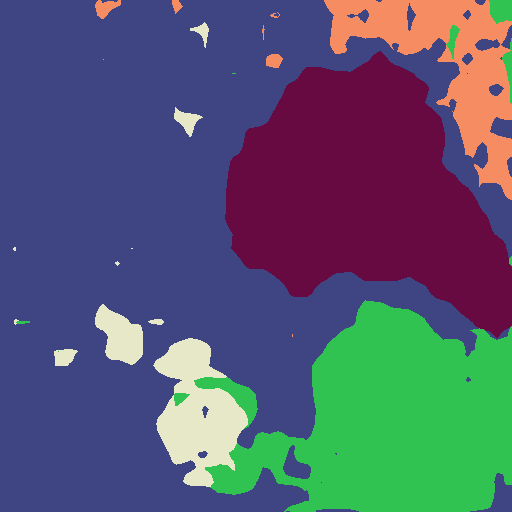}
    \end{subfigure}
    \begin{subfigure}{0.18\linewidth}
        \includegraphics[height=1.35cm]{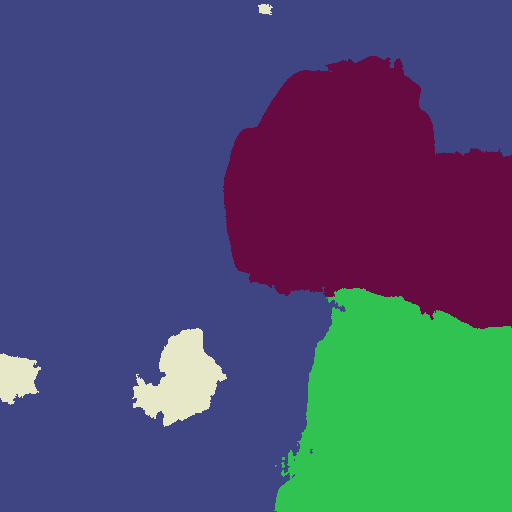}
    \end{subfigure}
    \begin{subfigure}{0.18\linewidth}
        \includegraphics[height=1.35cm]{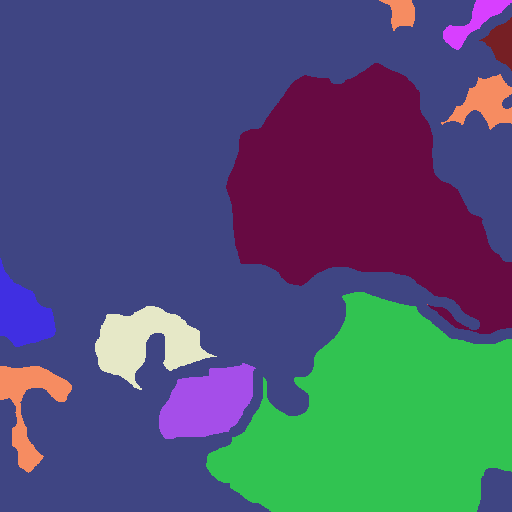}
    \end{subfigure}

    % Row 2
    \begin{subfigure}{0.18\linewidth}
        \includegraphics[height=1.35cm]{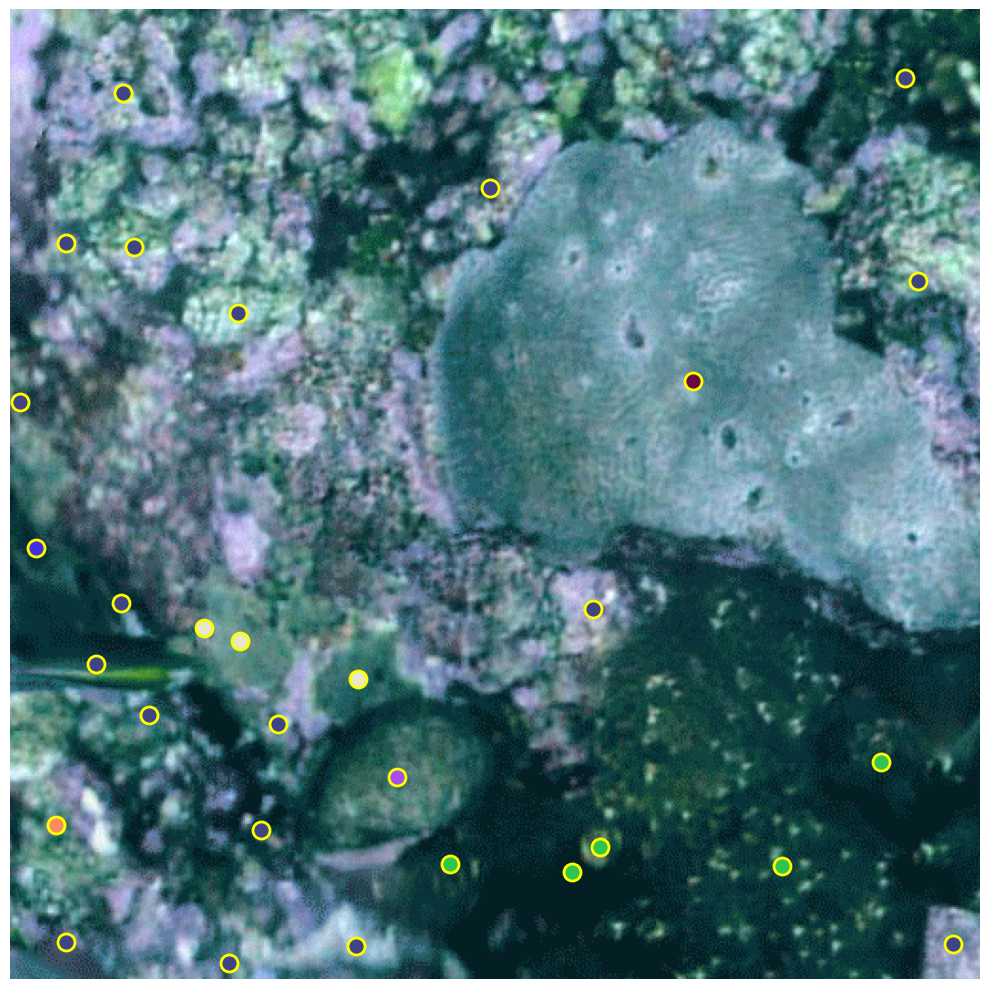}
    \end{subfigure}
    \begin{subfigure}{0.18\linewidth}
        \includegraphics[height=1.35cm]{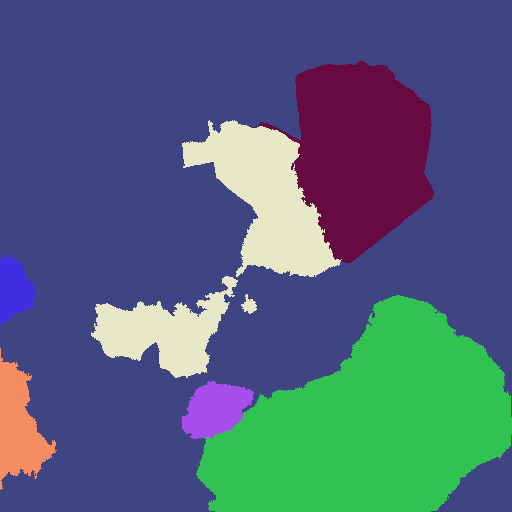}
    \end{subfigure}
    \begin{subfigure}{0.18\linewidth}
        \includegraphics[height=1.35cm]{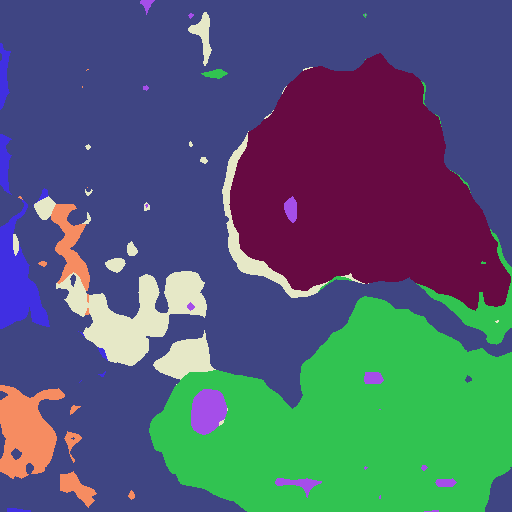}
    \end{subfigure}
    \begin{subfigure}{0.18\linewidth}
        \includegraphics[height=1.35cm]{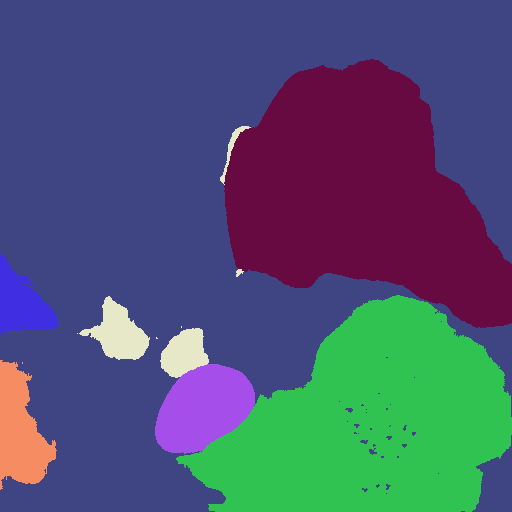}
    \end{subfigure}
    \begin{subfigure}{0.18\linewidth}
        \includegraphics[height=1.35cm]{images/MosaicsUCSD/gt_1.png}
    \end{subfigure}

    \vspace{2mm}

    % Row 3
    \begin{subfigure}{0.18\linewidth}
        \includegraphics[height=1.35cm]{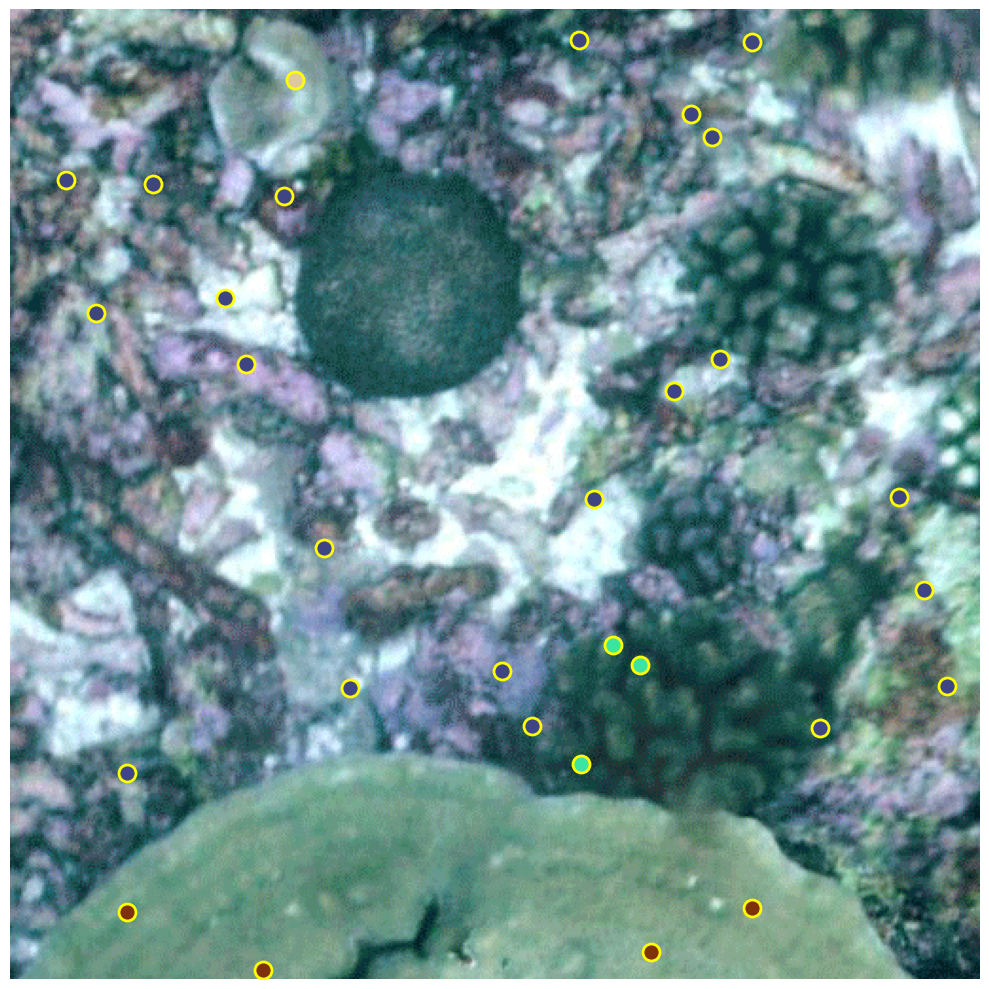}
    \end{subfigure}
    \begin{subfigure}{0.18\linewidth}
        \includegraphics[height=1.35cm]{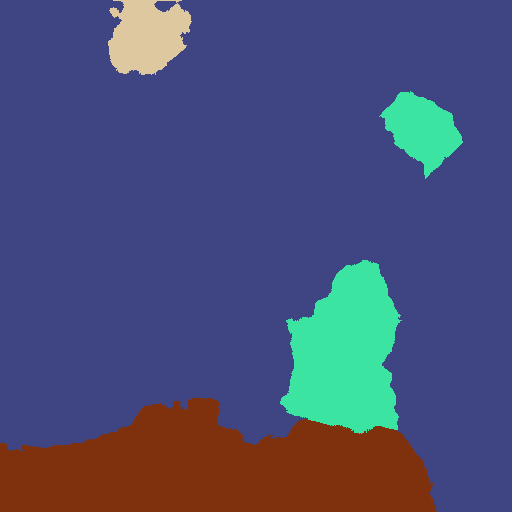}
    \end{subfigure}
    \begin{subfigure}{0.18\linewidth}
        \includegraphics[height=1.35cm]{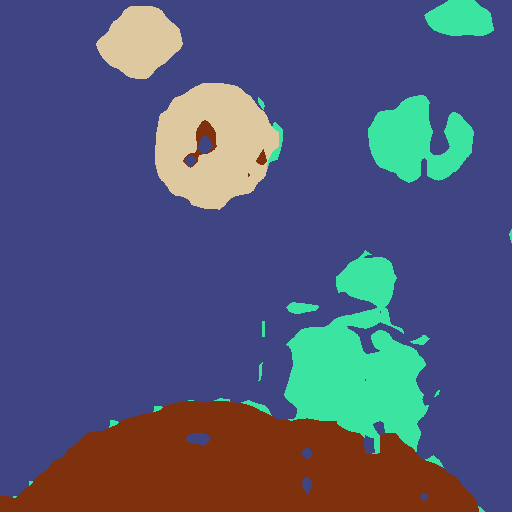}
    \end{subfigure}
    \begin{subfigure}{0.18\linewidth}
        \includegraphics[height=1.35cm]{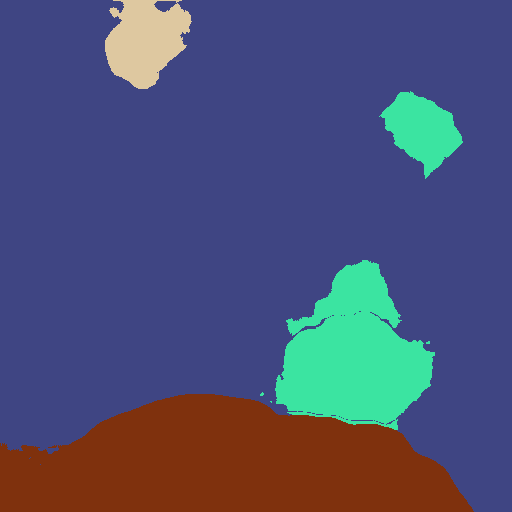}
    \end{subfigure}
    \begin{subfigure}{0.18\linewidth}
        \includegraphics[height=1.35cm]{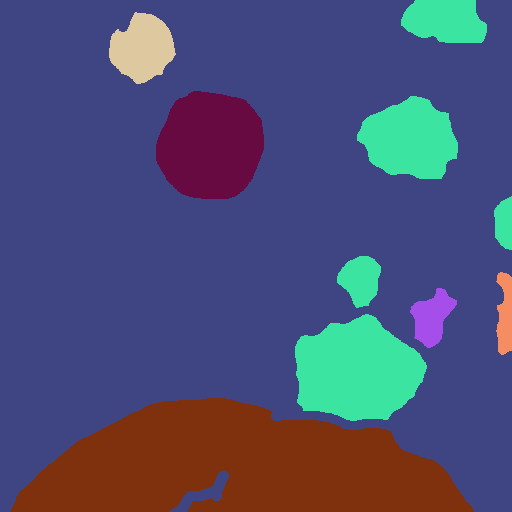}
    \end{subfigure}

    % Row 4
    \begin{subfigure}{0.18\linewidth}
        \includegraphics[height=1.35cm]{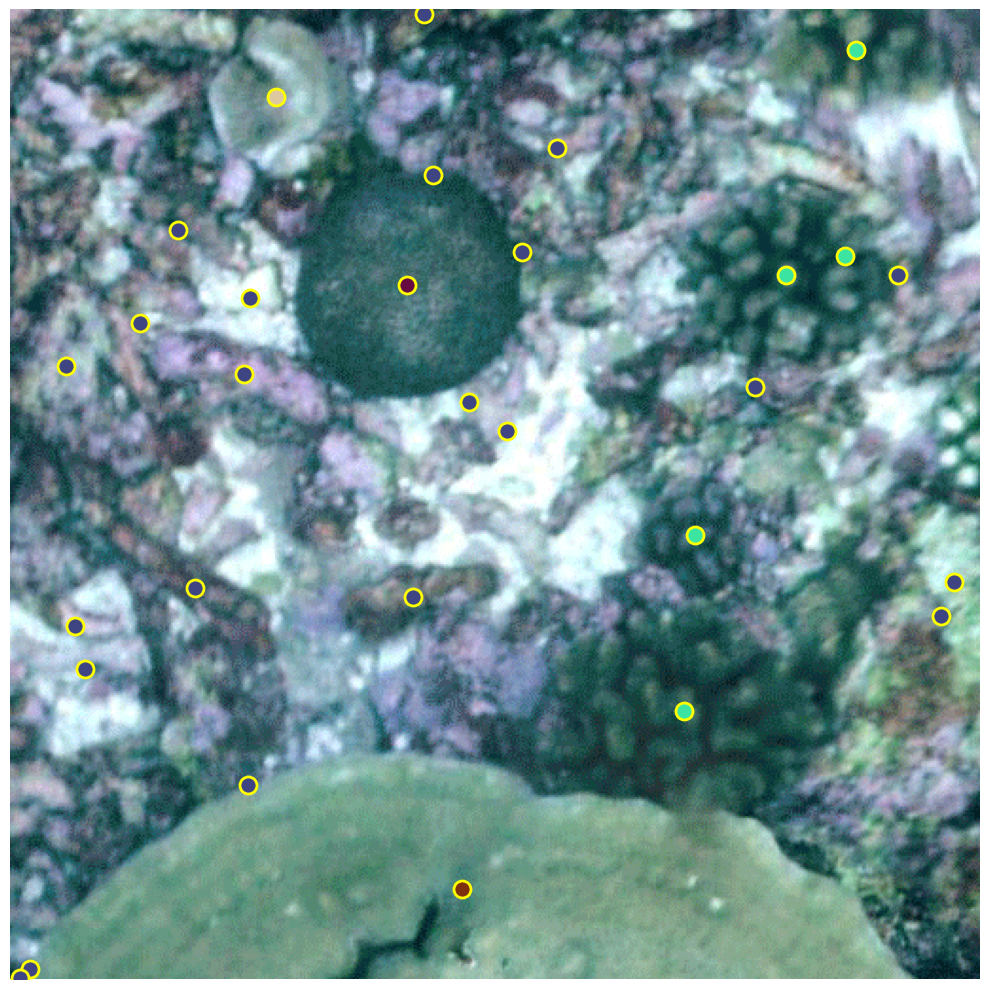}
    \end{subfigure}
    \begin{subfigure}{0.18\linewidth}
        \includegraphics[height=1.35cm]{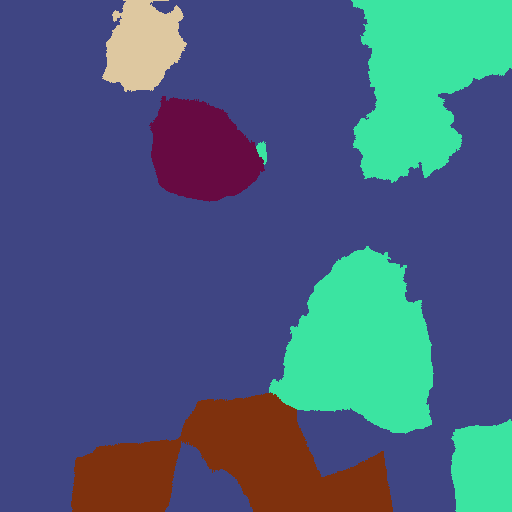}
    \end{subfigure}
    \begin{subfigure}{0.18\linewidth}
        \includegraphics[height=1.35cm]{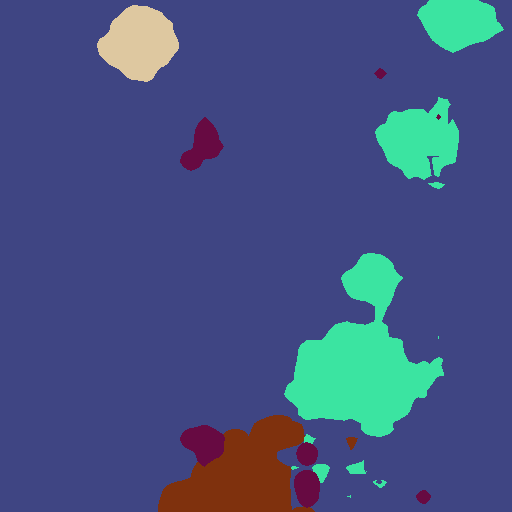}
    \end{subfigure}
    \begin{subfigure}{0.18\linewidth}
        \includegraphics[height=1.35cm]{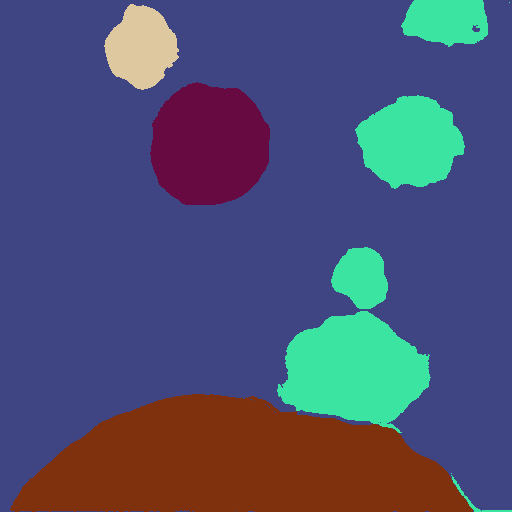}
    \end{subfigure}
    \begin{subfigure}{0.18\linewidth}
        \includegraphics[height=1.35cm]{images/MosaicsUCSD/gt_2.png}
    \end{subfigure}

    \vspace{2mm}

    % Row 5
    \begin{subfigure}{0.18\linewidth}
        \includegraphics[height=1.35cm]{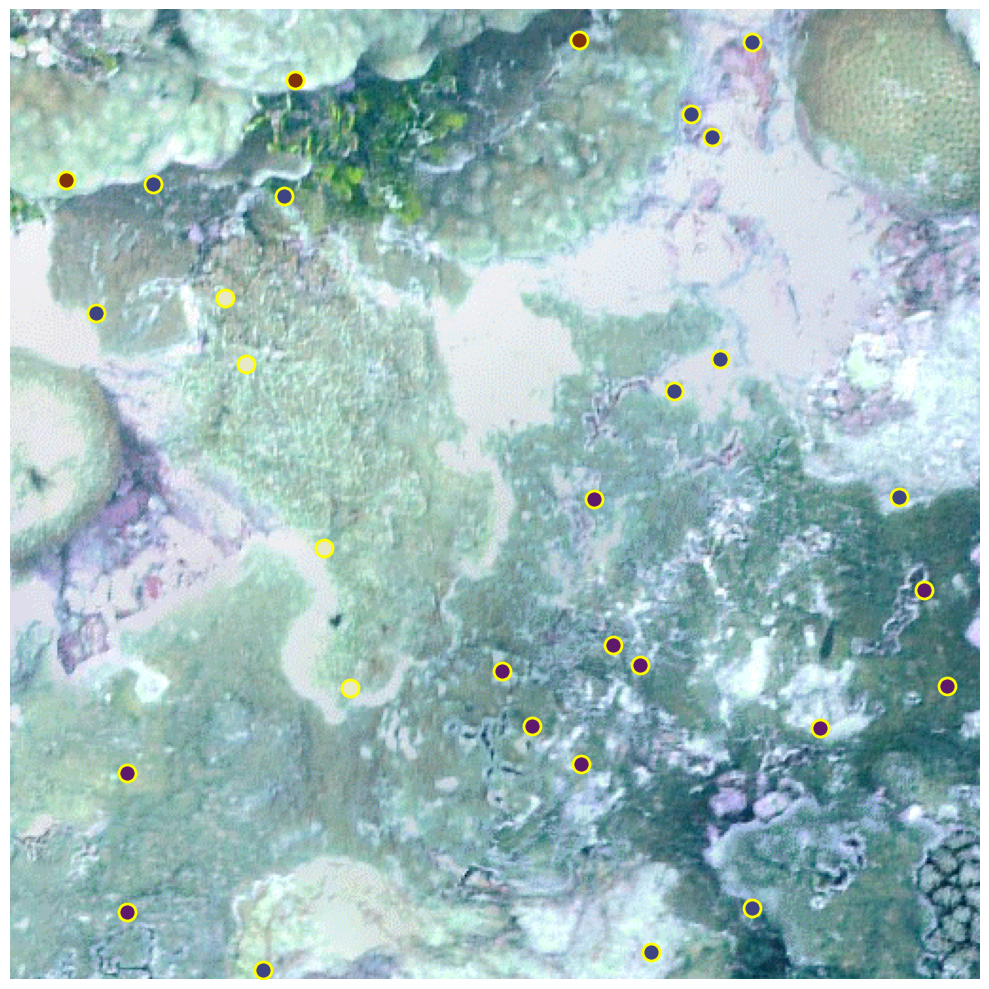}
    \end{subfigure}
    \begin{subfigure}{0.18\linewidth}
        \includegraphics[height=1.35cm]{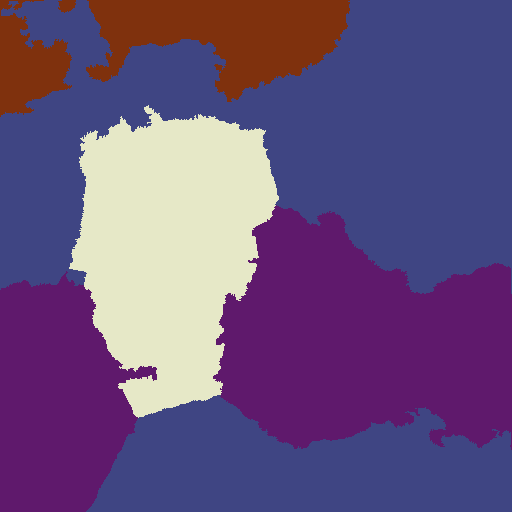}
    \end{subfigure}
    \begin{subfigure}{0.18\linewidth}
        \includegraphics[height=1.35cm]{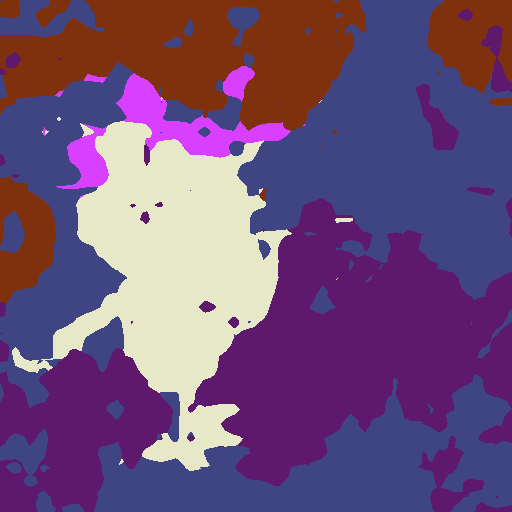}
    \end{subfigure}
    \begin{subfigure}{0.18\linewidth}
        \includegraphics[height=1.35cm]{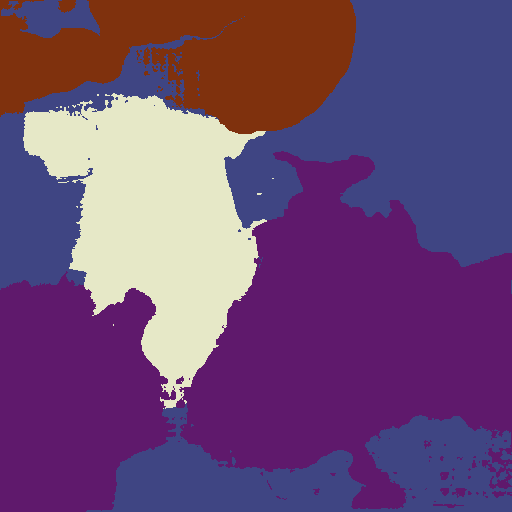}
    \end{subfigure}
    \begin{subfigure}{0.18\linewidth}
        \includegraphics[height=1.35cm]{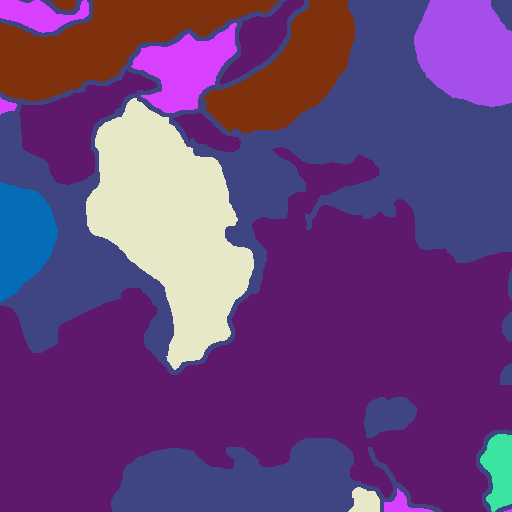}
    \end{subfigure}

    % Row 6
    \begin{subfigure}{0.18\linewidth}
        \includegraphics[height=1.35cm]{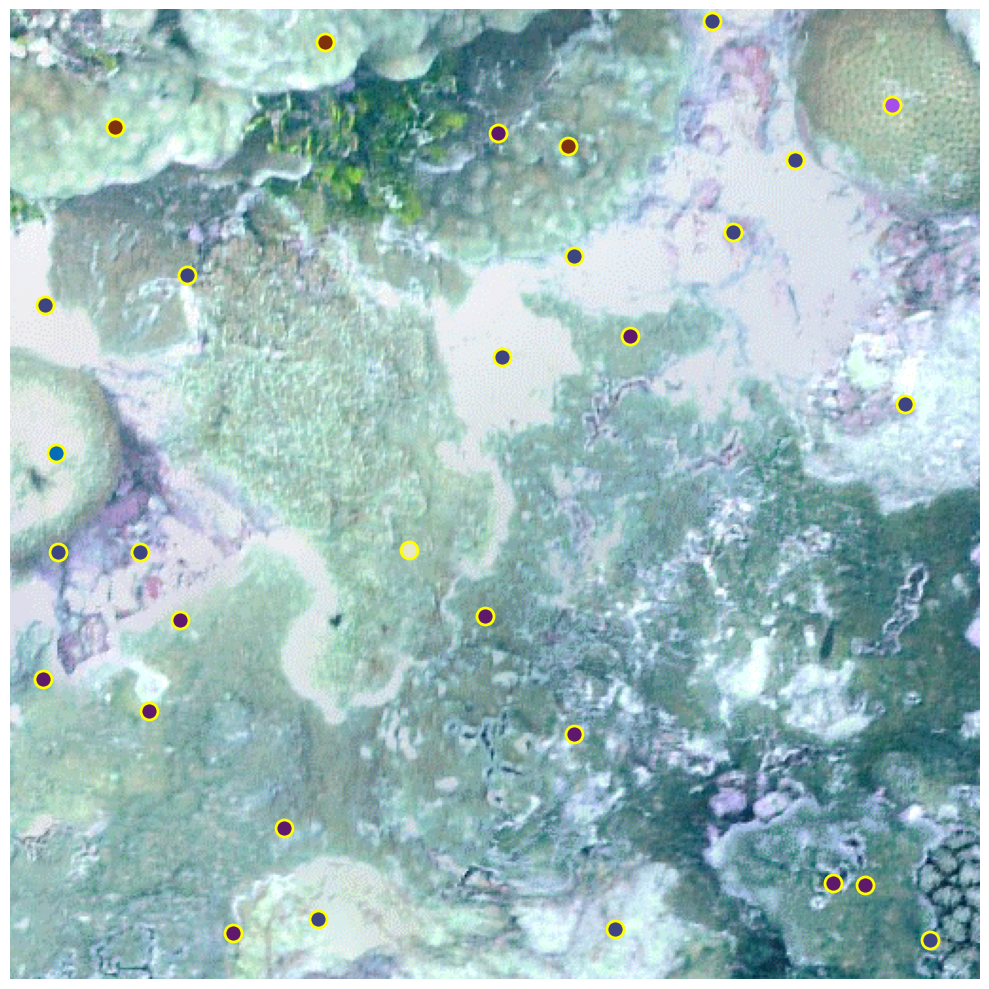}
        \caption*{\footnotesize (a) Img+PL}
    \end{subfigure}
    \begin{subfigure}{0.18\linewidth}
        \includegraphics[height=1.35cm]{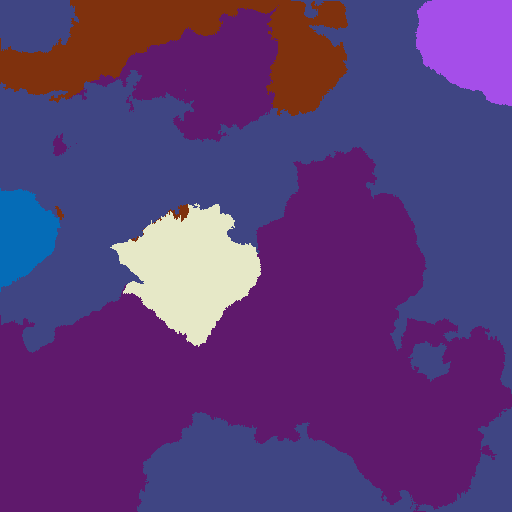}
        \caption*{\footnotesize (b) PLAS}
    \end{subfigure}
    \begin{subfigure}{0.18\linewidth}
        \includegraphics[height=1.35cm]{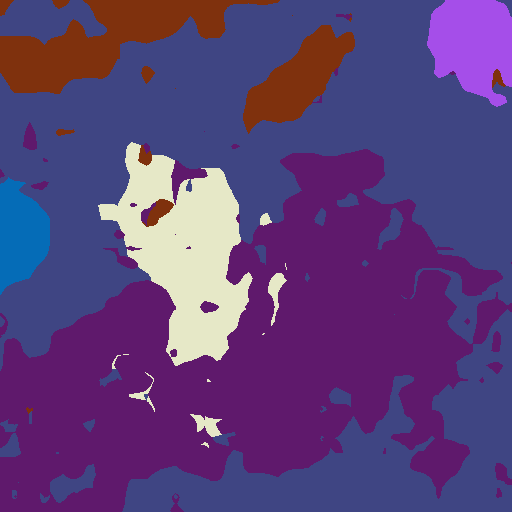}
        \caption*{\footnotesize (c) D+NN}
    \end{subfigure}
    \begin{subfigure}{0.18\linewidth}
        \includegraphics[height=1.35cm]{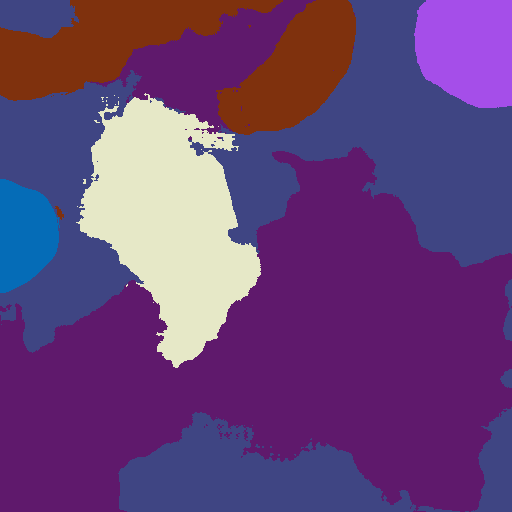}
        \caption*{\footnotesize (d) \textbf{Ours}}
    \end{subfigure}
    \begin{subfigure}{0.18\linewidth}
        \includegraphics[height=1.35cm]{images/MosaicsUCSD/gt_5.png}
        \caption*{\footnotesize (e) GT}
    \end{subfigure}

    \vspace{2mm}
    
    \caption{Qualitative results. Label augmentation using PLAS, D+NN and SSeg on UCSD Mosaics. Each pair of rows shows the final segmentation of the same image, with random sampling on top and our DynamicPoints sampling at the bottom. In each block, from left to right: (a) Input image with point-labels, (b) PLAS covers broad areas but with coarse region boundaries, (c) D+NN uses global feature matching but suffers from spatial incoherence and texture misclassification. (d) SSeg (Ours) merges SAM2's precision with PLAS's coverage to achieve accurate, coherent boundaries better matching the (e) Ground Truth. Our DynamicPoints sampling leads to a better point placement to support effective label augmentation. 
    %\fcolorbox{black}{background}{\rule{0pt}{2pt}\rule{3pt}{0pt}} Background color.
    }
\label{fig:sparse_sampling_multi}
\end{figure}

\subsubsection{Analysis of the label augmentation step}
%\paragraph{Experiment 2: Label augmentation evaluation.}
This experiment compares different label propagation strategies using a fixed set of randomly sampled points. To ensure the same set, we run all baselines locally on the same hardware and report results for the exact 25 and 300 random point sets  Table~\ref{tab:results_label_aug_ucsd} %and~\ref{tab:results_label_aug_skyscapes} 
presents a summary of the results with the two public benchmarks considered. 
On the UCSD Mosaics dataset, SSeg achieves the best results at both densities, surpassing D+NN and the specialized CoralSCOP model (e.g., reaching \textbf{86.04\%} masked-mIoU at 300 points compared to 82.69\% for D+NN).
The aerial SkyScapes dataset presents a different challenge. While SSeg remains competitive with D+NN at 300 points, D+NN proves slightly more robust at the sparse limit of 25 points (25.23\% vs. 23.68\% mIoU). 

Analyzing in more detail the different components of SSeg, note that across both domains, the full SSeg pipeline consistently outperforms the SSeg-SAM2Only baseline. This confirms that our PLAS-based filling module contributes to resolve unassigned regions that the foundation model misses.

%\textbf{Ablation: Overlap Resolution.}
As an additional ablation experiment, we evaluate the impact of our strategy to resolve conflicting masks during point propagation with the experiment summarized in Table~\ref{tab:overlap_solving_ablation}. Comparing our strategy against naive ``First-come'' and ``Last-come'' (if a point gets labeled by multiple expansions it keeps the label of the first one or the last, respectively) on SkyScapes reveals a big gap. The naive strategies keeps at $\approx$37\% mIoU, while our resolution algorithm reaches \textbf{45.21\%}. This proves that our logic for handling overlaps is a key factor for the final performance.

\begin{table}[!tb]
\caption{Comparison of \textbf{label augmentation approaches from fixed random point sets (25 or 300) based on superpixels or foundation model segmentations} on the \textbf{UCSD Mosaics} and \textbf{SkyScapes} datasets. All experiments ran in our environment using available public code for PLAS, D+NN and CoralSCOP. \textbf{Bold} and \underline{underlined} indicate the highest and second-highest, respectively. Computational times reported as mean ($\pm$std) across all images.}
\label{tab:results_label_aug_ucsd}
\centering
\resizebox{0.48\textwidth}{!}{
\begin{tabular}{lccccc}
\toprule
\multicolumn{6}{c}{\textbf{UCSD Mosaics} dataset}\\
\midrule

\multirow{2}{*}{Method} & \multirow{2}{*}{mPA} & \multirow{2}{*}{mIoU} & masked & masked & Time per \\
 &  &  & mPA & mIoU & Image (s) $\downarrow$ \\
\midrule

\multicolumn{6}{l}{\textbf{\textit{300 Sparse Points}}} \\
\midrule

PLAS-\textit{Ens}. & 83.87 & \underline{74.27} & 83.75 & 81.41 & \underline{1.22} ($\pm$0.1) \\
CoralSCOP & 67.70 & 52.57 & 67.82 & 59.74 & \textbf{0.71} ($\pm$0.1) \\
D+NN & \underline{85.03} & \underline{74.57} & \underline{84.99} & \underline{82.69} & 4.50 ($\pm$0.1) \\
SSeg-SAM2Only (Ours) & 84.55 & 74.09 & 84.49 & 82.51 & 5.14 ($\pm$1.7) \\
SSeg (Ours) & \textbf{88.44} & \textbf{75.40} & \textbf{88.59} & \textbf{86.04} & 6.18 ($\pm$1.5) \\

\midrule
\multicolumn{6}{l}{\textbf{\textit{25 Sparse Points}}} \\
\midrule

PLAS-\textit{Ens}. & 61.27 & 44.78 & 60.77 & 52.55 & 1.14 ($\pm$0.1) \\
CoralSCOP & 55.61 & 45.25 & 54.99 & 49.58 & \textbf{0.11} ($\pm$0.1) \\
D+NN & \underline{66.16} & \textbf{51.85} & \underline{65.80} & \underline{59.11} & 4.48 ($\pm$0.1) \\
SSeg-SAM2Only (Ours) & 53.96 & 49.74 & 52.00 & 51.11 & \underline{0.67} ($\pm$0.1) \\
SSeg (Ours) & \textbf{67.73} & \underline{50.55} & \textbf{67.41} & \textbf{59.85} & 2.23 ($\pm$0.3) \\

\bottomrule
\end{tabular}
}

\centering
\resizebox{0.5\textwidth}{!}{
\begin{tabular}{lccc}

\toprule
\multicolumn{4}{c}{\textbf{SkyScapes} dataset}\\
\midrule
Method & mPA ($\pm$std) & mIoU ($\pm$std) & Time per Img (s) $\downarrow$ \\
\midrule

\multicolumn{4}{l}{\textbf{\textit{300 Sparse Points}}} \\
\midrule

PLAS-\textit{Ens}. & 52.12 ($\pm$25.7) & 41.67 ($\pm$22.6) & \underline{3.92} ($\pm$0.8) \\

D+NN & \underline{54.14} ($\pm$21.6) & 40.44 ($\pm$21.2) & \textbf{3.04} ($\pm$0.1) \\

SSeg-SAM2Only (Ours) & 52.11 ($\pm$22.1) & \underline{43.00} ($\pm$21.9) & 10.67 ($\pm$1.7) \\
SSeg (Ours) & \textbf{55.38} ($\pm$23.1) & \textbf{43.91} ($\pm$22.2) & 12.38 ($\pm$1.7) \\

\midrule
\multicolumn{4}{l}{\textbf{\textit{25 Sparse Points}}} \\
\midrule

PLAS-\textit{Ens}. & 29.27 ($\pm$23.4) & 20.54 ($\pm$17.2) & 3.65 ($\pm$0.7) \\

D+NN & \textbf{36.03} ($\pm$23.1) & \textbf{25.23} ($\pm$19.2) & \underline{3.06} ($\pm$0.1) \\

SSeg-SAM2Only (Ours) & 23.56 ($\pm$19.8) & 21.21 ($\pm$18.3) & \textbf{2.83} ($\pm$0.7) \\
SSeg (Ours) & \underline{32.46} ($\pm$25.3) & \underline{23.68} ($\pm$19.1) & 4.17 ($\pm$0.7) \\

\bottomrule
\end{tabular}
}
\end{table}

\begin{table}[!htb]
\centering
\caption{\textbf{Impact of overlap resolution strategy} with the SkyScapes dataset. We compare our proposed strategy to resolve overlapping conflicts during propagation of 300 grid points against naive heuristics (keeping the `First' or `Last' generated mask). Our approach significantly reduces conflict errors, boosting both metrics by over 7\%.}
\label{tab:overlap_solving_ablation}
\footnotesize
\begin{tabular}{l c c | c c}
\toprule
\textbf{Resolution Strategy} & \textbf{mIoU} & \textbf{$\Delta$} & \textbf{mPA} & \textbf{$\Delta$} \\
\midrule
First-come & 36.67 & -- & 46.99 & -- \\
Last-come & 37.80 & +1.13 & 48.29 & +1.30 \\
\midrule
\textbf{SSeg (Ours)} & \textbf{45.21} & \textbf{+7.41} & \textbf{56.01} & \textbf{+7.72} \\
\bottomrule
\end{tabular}

\end{table}

\begin{table*}[!tb]
\caption{\textbf{Sparse point selection methods} Performance of \textbf{SSeg} for 25 point-labels in UCSD Mosaics and SkyScapes using different static or active (dynamic) point sampling strategies. \textit{Centroid SAM2-guided} selects the centroids of the largest automatic SAM2 masks. \textit{DynamicPoints-onlyA} selects 100\% of points using our acquisition function $A(p)$, while \textit{DynamicPoints} combines 50\% $A(p)$ selection with 50\% random sampling. \textbf{Bold} and \underline{underlined} means highest and second highest, respectively. Computational times are given as mean ($\pm$std) for all images.}
\footnotesize
\label{tab:results_point_sel}
\centering
\resizebox{0.8\textwidth}{!}{
\begin{tabular}{lccc|ccc}
\toprule
\multirow{2}{*}{Sampling strategy} & \multicolumn{3}{c}{UCSD Mosaics} & \multicolumn{3}{c}{Skyscapes} \\
\cmidrule(lr){2-4} \cmidrule(lr){5-7}
 & mPA & mIoU & Time per Image (s) $\downarrow$ & mPA & mIoU & Time per Image (s) $\downarrow$ \\
\midrule

\textbf{\textit{Static point sampling strategies }}\\
\midrule
Random & 67.73 & 50.55 & \underline{2.2} ($\pm$0.29) & 32.46 & 23.68 & \textbf{4.3} ($\pm$0.60) \\
Grid & \underline{74.43} & \underline{58.44} & \textbf{2.2} ($\pm$0.17) & 32.03 & 23.02 & 4.3 ($\pm$0.60) \\
\midrule
\textbf{\textit{Dynamic point sampling strategies}}\\
\midrule
Centroid SAM2-guided & 71.85 & 49.93 & 2.7 ($\pm$0.37) & \underline{40.55} & 23.86 & 4.3 ($\pm$0.71) \\

DynamicPoints-onlyA (Ours) & 73.19 & 51.18 & 3.7 ($\pm$0.65) & \textbf{40.87} & \underline{24.50} & 6.4 ($\pm$1.49) \\

DynamicPoints (Ours) & \textbf{76.59} & \textbf{59.91} & 3.4 ($\pm$0.57) & 36.51 & \textbf{27.32} & 5.9 ($\pm$1.27) \\
\bottomrule
\end{tabular}
}
\end{table*}

\subsubsection{Analysis of the sparse point selection step.}
This final experiment analyzes more exhaustively the impact of the strategies to select the sparse points location. We fix the label propagation model to the best performing (SSeg) and change only the strategy used to select the points. Table~\ref{tab:results_point_sel} shows the quality of the resulting dense annotation obtained after propagation when selecting 25 points with different strategies. Our \textit{DynamicPoints} strategy achieves the best results overall.
It is worth noting that grid sampling performs surprisingly well on UCSD (58.44\%), likely due to the dense, texture-like distribution of coral species which benefits from uniform spatial coverage. However, grid sampling fails to generalize to the aerial domain (SkyScapes), where it performs worse than random  (23.02\% vs 23.68\%). In contrast, \textit{DynamicPoints} demonstrates superior robustness, consistently outperforming all static and dynamic baselines across both domains.
Interestingly, simply picking the centers of the largest objects (\textit{Centroid SAM2-guided}) significantly degrades the performance (49.93\%), highlighting the significance of exploring points within background regions.

To understand why our strategy works best, we analyzed two key components (Table~\ref{tab:ablations_combined}). First, the mix of random points: using only points obtained by A(p) (0\% random) gives lower results (51.18\% mIoU). The performance peaks when we use 50\% random points (\textbf{59.91\%}). This confirms that random points are necessary to cover low-contrast background areas that the model misses. Second, the parameter $\lambda$: the best result is found at $\lambda=0.5$, indicating that we need an equal balance between finding object centers (exploitation) and exploring empty space (exploration).

\begin{table}[!tb]
\centering
\caption{\textbf{SSeg component sensitivity analysis.} Impact of the acquisition balance $\lambda$ (left) and the random sampling ratio (right) on the performance of SSeg on the UCSD Mosaics dataset (propagating 30 points). For the left block, the random ratio is fixed at 50\%. For the right block, $\lambda$ is fixed at 0.5. A balanced configuration yields optimal performance in both.}
\label{tab:ablations_combined}
\resizebox{0.9\columnwidth}{!}{
\footnotesize
\begin{tabular}{c cc | c cc}
\toprule
\multicolumn{3}{c|}{\textbf{Acquisition Balance ($\lambda$)}} & \multicolumn{3}{c}{\textbf{Random Sampling Ratio}} \\
\cmidrule(r){1-3} \cmidrule(l){4-6}
\textbf{$\lambda$} & \textbf{mIoU} & \textbf{mPA} & \textbf{Ratio} & \textbf{mIoU} & \textbf{mPA} \\
\midrule
0.00 & 57.77 & 71.52 & 0\% & 54.61 & 74.25 \\
0.25 & 62.73 & 78.74 & 25\% & 61.15 & 78.84 \\
\textbf{0.50} & \textbf{63.33} & 79.55 & \textbf{50\%} & \textbf{63.33} & \textbf{79.55} \\
0.75 & 63.25 & \textbf{79.60} & 75\% & 58.37 & 72.30 \\
1.00 & 62.41 & 78.50 & 100\% & 54.81 & 70.43 \\
\bottomrule
\end{tabular}
}
\end{table}

\begin{table}[!htb]
\caption{\textbf{Downstream task training effectiveness.} SegFormer performance on UCSD Mosaics using Dense GT vs. sparse propagations (25 points), with sampling strategies indicated in the label source. Note that D+NN utilizes `Oracle-based' guidance for the first 10 points. Despite this privilege, SSeg outperforms all baselines, recovering \textbf{90\%} of the fully supervised performance with \textbf{four orders of magnitude} fewer annotations.}
\label{tab:results_train}
\centering
\resizebox{0.9\columnwidth}{!}{
\footnotesize
\begin{tabular}{l c c c}
\toprule
\textbf{Label Source} & \textbf{Points / Img $\downarrow$} & \textbf{mPA}$\uparrow$ & \textbf{mIoU}$\uparrow$ \\
\midrule
Dense GT & $262{,}144$ & \textbf{68.51} & \textbf{57.87} \\
\midrule
PLAS-Random & 25 & 45.79 & 35.39 \\
PLAS-DynamicPoints & 25 & 53.80 & 41.02 \\
D+NN-Active~\cite{raine2024human} & 25 & 60.61 & 46.85 \\
\textbf{SSeg (Ours)} & 25 & \underline{61.66} & \underline{49.15} \\
\bottomrule
\end{tabular}
}
\end{table}

\subsubsection{Utility for downstream training.}
Beyond expert analysis, we verify if SSeg masks are sufficient to train automated models. To test this, we fine-tune a SegFormer-B2~\cite{xie2021segformer} model (pre-trained on ADE20K~\cite{zhou2017scene}) on UCSD Mosaics. We train for 80 epochs using the AdamW optimizer ($lr=10^{-4}$) with the 25-point masks from different methods and compare them against training with full Ground Truth (GT). As shown in Table~\ref{tab:results_train}, applying DynamicPoints to PLAS improves scores by over +8\% compared to random sampling. Notably, SSeg outperforms D+NN without requiring "Oracle-based" initialization. Remarkably, using only 25 points is a \textbf{99.99\% reduction} in labels compared to the full 262,144 pixels. Despite a 99.99\% reduction in labels (25 points vs. 262,144 pixels), SSeg reaches 90\% of fully supervised performance (61.66\% vs. 68.51\% mIoU), proving sparse clicks can effectively train standard segmentation models.

\section{Conclusions}
This work addresses the challenge of generating dense semantic segmentations in complex unstructured environments, such as underwater and aerial domains, from scarce and costly expert annotations. We introduce SSeg, a framework that couples an active point selection method (DynamicPoints) with a hybrid label propagation strategy. By merging SAM2’s precise boundaries with the coverage of superpixel methods, SSeg mitigates spatial incoherence typical of feature-based propagation. Experiments on UCSD Mosaics and SkyScapes confirm state-of-the-art performance, where SSeg consistently surpasses baselines across different sampling strategies. Crucially, SSeg uses an \textbf{automatic point selection} method, avoiding the need for the manual `Oracle-based' initialization required by D+NN, while our overlap resolution strategy ensures robust handling of dense object clusters. Furthermore, we demonstrate the practical utility of our approach for downstream tasks, showing that SSeg-propagated masks can recover nearly the full performance of dense supervision while reducing annotation effort by four orders of magnitude. SSeg provides a scalable workflow to minimize annotation effort in marine ecology. Future work will integrate deep spatial feature similarity, combining precise spatial details with semantic understanding to further improve label quality with limited resources.

\section{Acknowledgments}
This work was supported by a DGA scholarship and by DGA project T45\_23R, and grants AIA2025-163563-C31, PID2024-159284NB-I00, PID2021-125514NB-I00 and PID2024-158322OB-I00 funded by MCIN/AEI/10.13039/501100011033 and ERDF.

%%%%%%%%% REFERENCES
{\small
\bibliographystyle{ieee_fullname}
\bibliography{egbib}
}

\clearpage
\setcounter{page}{1}
\maketitlesupplementary

\begin{strip}
    \centering
    
    % --- TABLE 1: Per-Class Performance ---
    \captionof{table}{\textbf{Per-class performance (SSeg, 25 active points).} Detailed mPA and mIoU breakdown for SkyScapes (left) and UCSD Mosaics (right). Data is split into multiple columns for compactness. Note that while UCSD Mosaics results are stable, SkyScapes shows high variation between classes. This is because thin objects (e.g., road lines) are very difficult to capture with sparse point-labels.}
    \label{tab:per_class_combined}
    
    \vspace{0.2cm}

    % --- Left Side: SkyScapes (2 Columns) ---
    \begin{minipage}[t]{0.38\textwidth}
        \centering
        \subcaption*{\textbf{(a) SkyScapes}}
        \resizebox{\linewidth}{!}{
            \begin{tabular}{lcc|lcc}
            \toprule
            \textbf{C} & \textbf{mPA} & \textbf{mIoU} & \textbf{C} & \textbf{mPA} & \textbf{mIoU} \\
            \midrule
            0  & 32.46 & 23.68 & 11 & 67.06 & 52.26 \\
            1  & 71.78 & 54.04 & 12 & 6.04  & 3.92  \\
            2  & 61.50 & 39.82 & 13 & 7.82  & 7.48  \\
            3  & 42.70 & 28.19 & 14 & 4.44  & 2.81  \\
            4  & 36.03 & 21.68 & 15 & 11.32 & 8.30  \\
            5  & 82.43 & 63.44 & 16 & 4.04  & 3.95  \\
            6  & 22.44 & 11.95 & 17 & 10.38 & 7.44  \\
            7  & 36.86 & 23.36 & 18 & 56.88 & 39.97 \\
            8  & 8.43  & 4.29  & 19 & 45.83 & 31.36 \\
            9  & 3.94  & 3.83  & 20 & 65.75 & 50.55 \\
            10 & 3.97  & 2.74  &    &       &       \\
            \bottomrule
            \end{tabular}
        }
    \end{minipage}
    \hfill
    % --- Right Side: UCSD Mosaics (3 Columns) ---
    \begin{minipage}[t]{0.58\textwidth}
        \centering
        \subcaption*{\textbf{(b) UCSD Mosaics}}
        \resizebox{\linewidth}{!}{
            \begin{tabular}{lcc|lcc|lcc}
            \toprule
            \textbf{C} & \textbf{mPA} & \textbf{mIoU} & \textbf{C} & \textbf{mPA} & \textbf{mIoU} & \textbf{C} & \textbf{mPA} & \textbf{mIoU} \\
            \midrule
            0 & 64.85 & 56.95 & 12 & 78.95 & 63.37 & 24 & 82.97 & 67.42 \\
            1 & 82.82 & 59.67 & 13 & 98.64 & 85.37 & 25 & 76.04 & 56.42 \\
            2 & 91.14 & 80.07 & 14 & 87.97 & 73.82 & 26 & 75.14 & 58.71 \\
            3 & 95.85 & 80.15 & 15 & 82.75 & 65.86 & 27 & 86.11 & 73.29 \\
            4 & 69.34 & 45.81 & 16 & 65.29 & 41.79 & 28 & 91.40 & 75.53 \\
            5 & 85.58 & 71.40 & 17 & 85.09 & 69.03 & 29 & 48.72 & 28.86 \\
            6 & 60.86 & 39.68 & 18 & 69.81 & 45.33 & 30 & 92.73 & 82.75 \\
            7 & 76.82 & 58.98 & 19 & 77.41 & 62.07 & 31 & 74.14 & 60.03 \\
            8 & 74.15 & 52.96 & 20 & 84.89 & 71.19 & 32 & 71.45 & 54.74 \\
            9 & 66.02 & 46.44 & 21 & 61.47 & 35.87 & 33 & 67.69 & 46.98 \\
            10 & 77.43 & 61.49 & 22 & 60.09 & 43.17 & Bg & 75.80 & 65.12 \\
            11 & 70.94 & 40.52 & 23 & 58.49 & 38.96 & & & \\
            \bottomrule
            \end{tabular}
        }
    \end{minipage}

    \vspace{1cm} % Gap between the two big tables

    % --- TABLE 2: Experiment 1 Results ---
    \captionof{table}{\textbf{Quantitative results from Experiment 1 (Numerical).} We report the masked-mIoU values corresponding to the plot in Figure 1 of the main paper.}
    \label{tab:results_relatedwork}
    \centering
    \resizebox{0.6\textwidth}{!}{
        \begin{tabular}{ll|cccc}
        \toprule
        \textbf{Method} & \textbf{Label} & \multicolumn{4}{c}{\textbf{masked-mIoU}}\\
         & \textbf{Style} & 5 & 10 & 25 & 300\\
        \midrule
        PLAS - \textit{Ens.}~\cite{raine2022point} & Rand.  & 25.91 & 35.56 & 50.46 & 85.45 \\
        D+NN~\cite{raine2024human} & Rand.  & 32.09 & 42.79 & 58.04 & 81.15 \\
        SparseUWSeg (Ours) & Rand.$^*$  & 35.27 & 42.73 & 62.90 & 86.04 \\
        PLAS - \textit{Ens.}~\cite{raine2022point} & Grid  & 30.00 & 40.34 & 59.82 & \underline{89.38} \\
        D+NN~\cite{raine2024human} & Grid & 35.74 & 50.58 & 64.40 & 85.77 \\
        SparseUWSeg (Ours) & Grid & \underline{38.44} & 51.70 & 67.30 & 86.31 \\
        D+NN~\cite{raine2024human} & Active & \textbf{52.60} & \textbf{59.48} & \underline{67.97} & 85.00 \\
        SparseUWSeg (Ours) & Active & 38.41 & \underline{53.76} & \textbf{70.40} & \textbf{93.15}  \\
        \bottomrule
        \multicolumn{6}{l}{* Random locations differ as baseline coordinates are unavailable.}%\\ 
        \end{tabular}
    }
\end{strip}

% \section{Additional qualitative results}

\begin{figure*}[!thb]
    \centering
    % Row 1
    \begin{subfigure}{0.18\linewidth}
        \includegraphics[width=\linewidth]{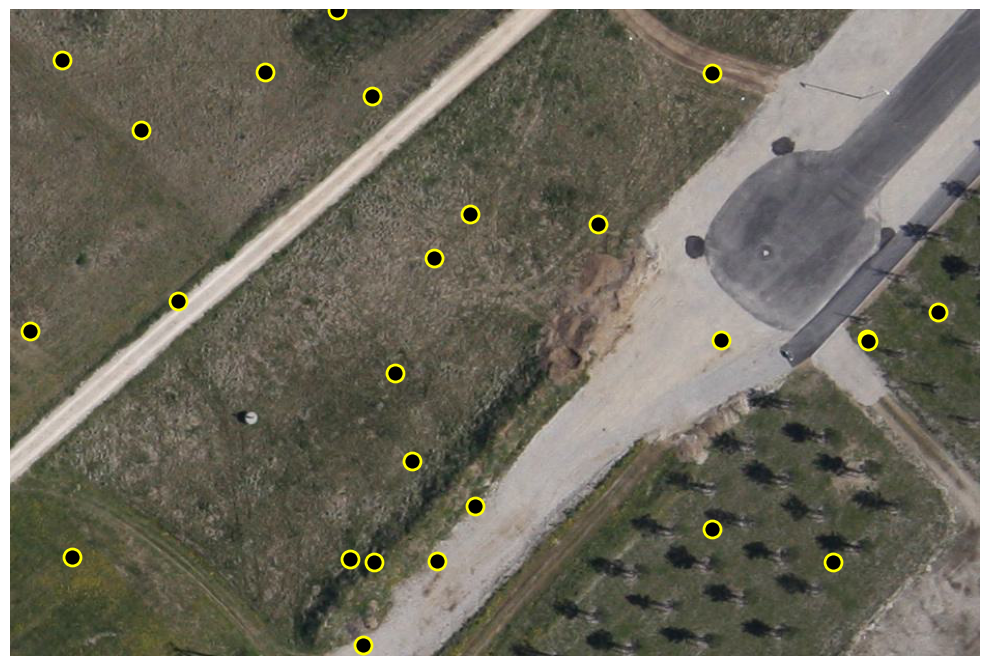}
    \end{subfigure}
    \begin{subfigure}{0.18\linewidth}
        \includegraphics[width=\linewidth]{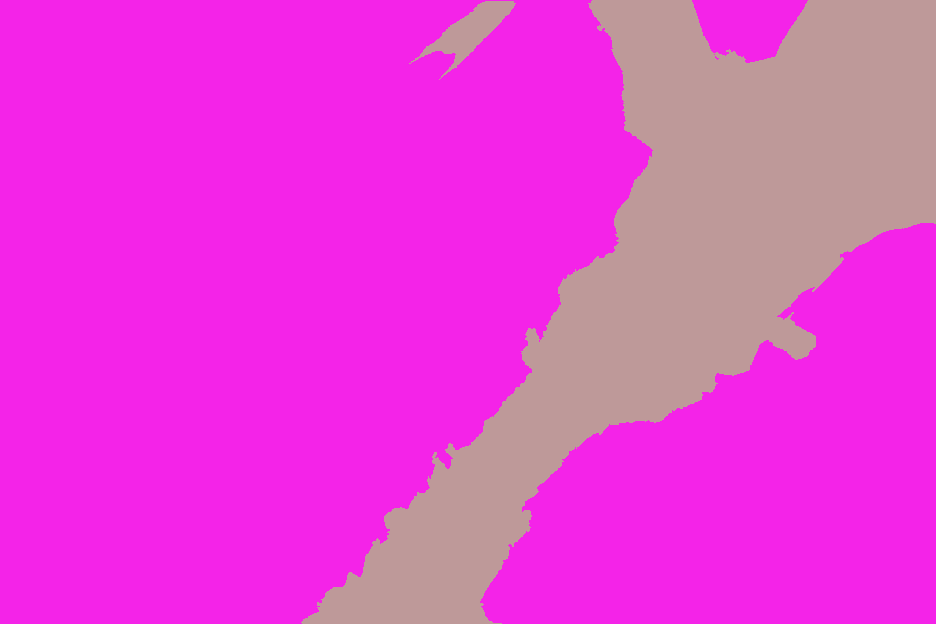}
    \end{subfigure}
    \begin{subfigure}{0.18\linewidth}
        \includegraphics[width=\linewidth]{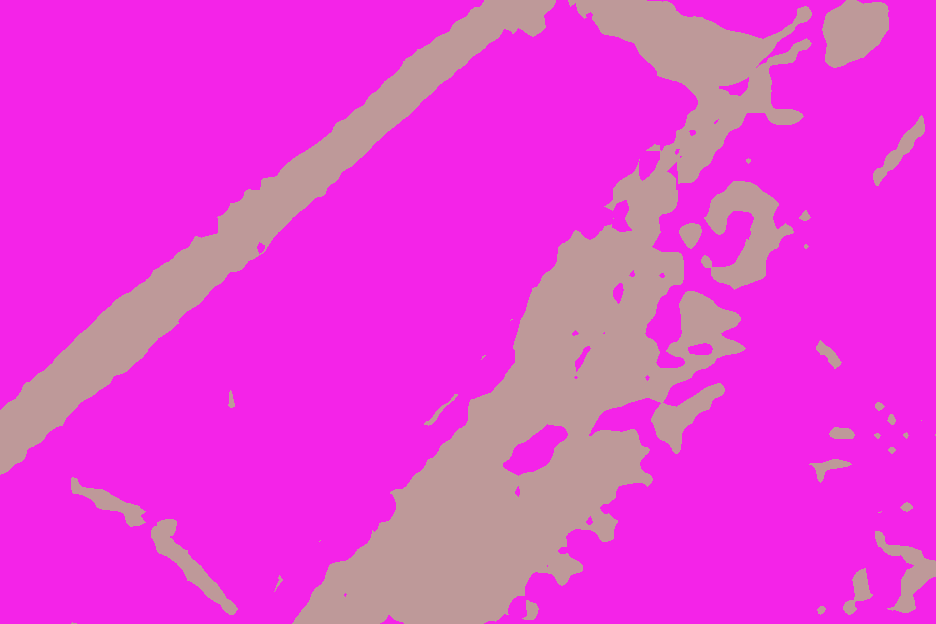}
    \end{subfigure}
    \begin{subfigure}{0.18\linewidth}
        \includegraphics[width=\linewidth]{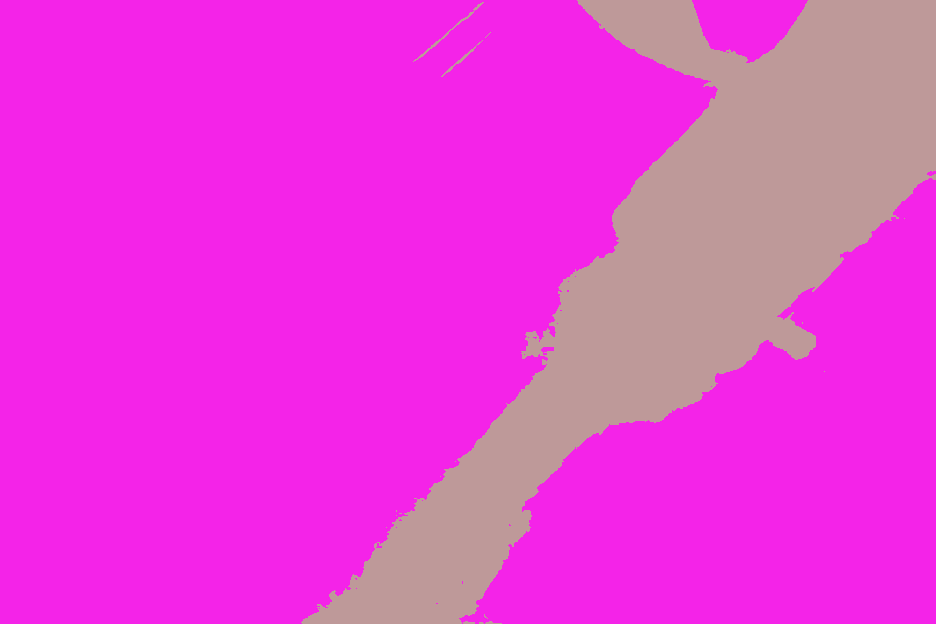}
    \end{subfigure}
    \begin{subfigure}{0.18\linewidth}
        \includegraphics[width=\linewidth]{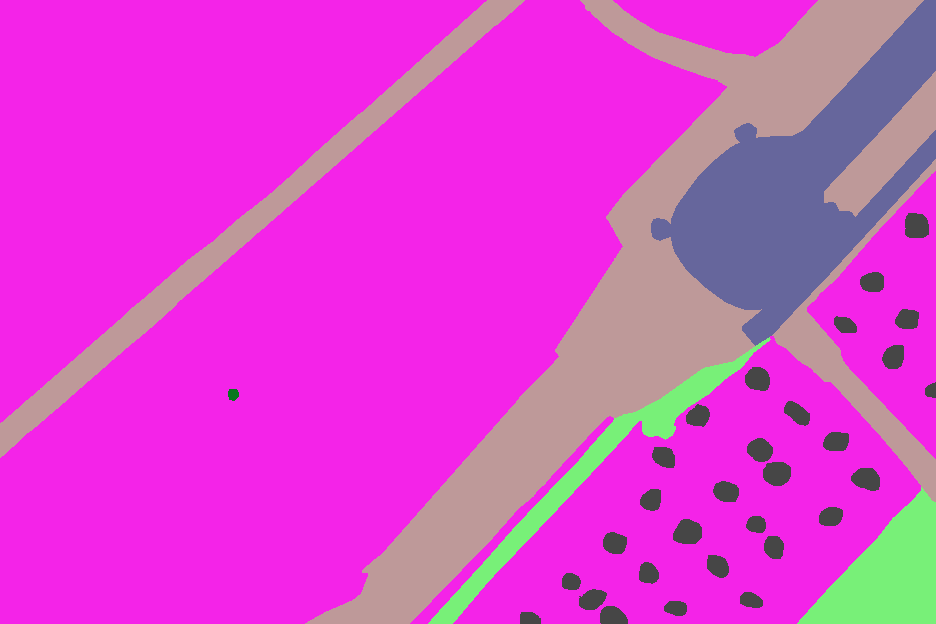}
    \end{subfigure}

    % Row 2
    \begin{subfigure}{0.18\linewidth}
        \includegraphics[width=\linewidth]{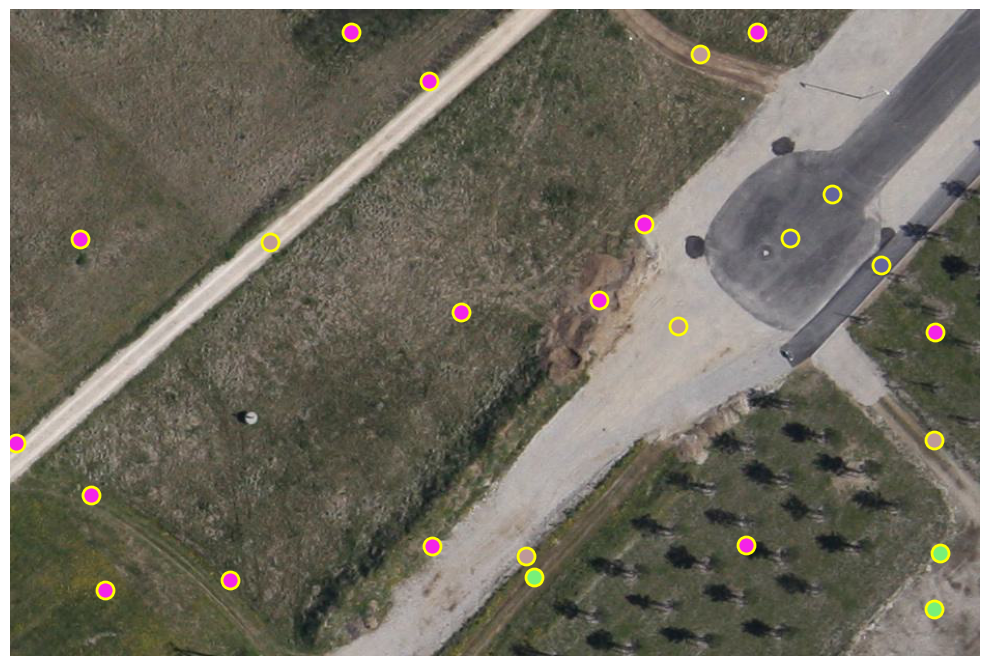}
    \end{subfigure}
    \begin{subfigure}{0.18\linewidth}
        \includegraphics[width=\linewidth]{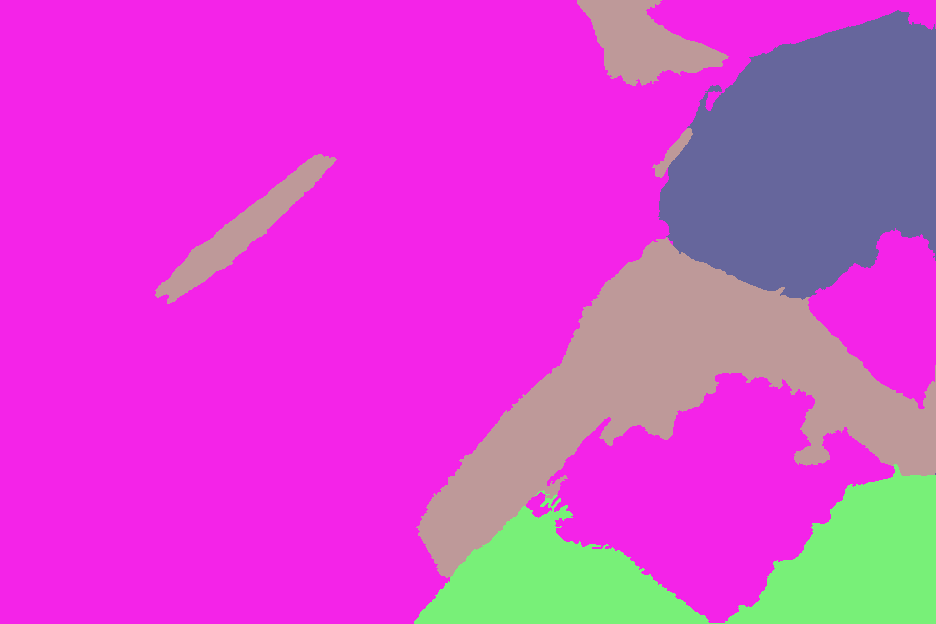}
    \end{subfigure}
    \begin{subfigure}{0.18\linewidth}
        \includegraphics[width=\linewidth]{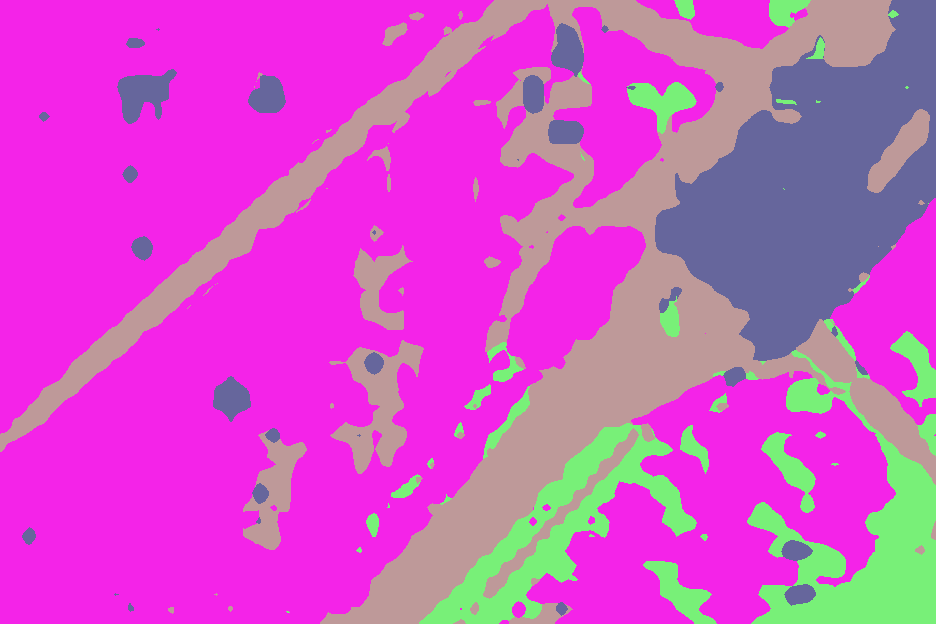}
    \end{subfigure}
    \begin{subfigure}{0.18\linewidth}
        \includegraphics[width=\linewidth]{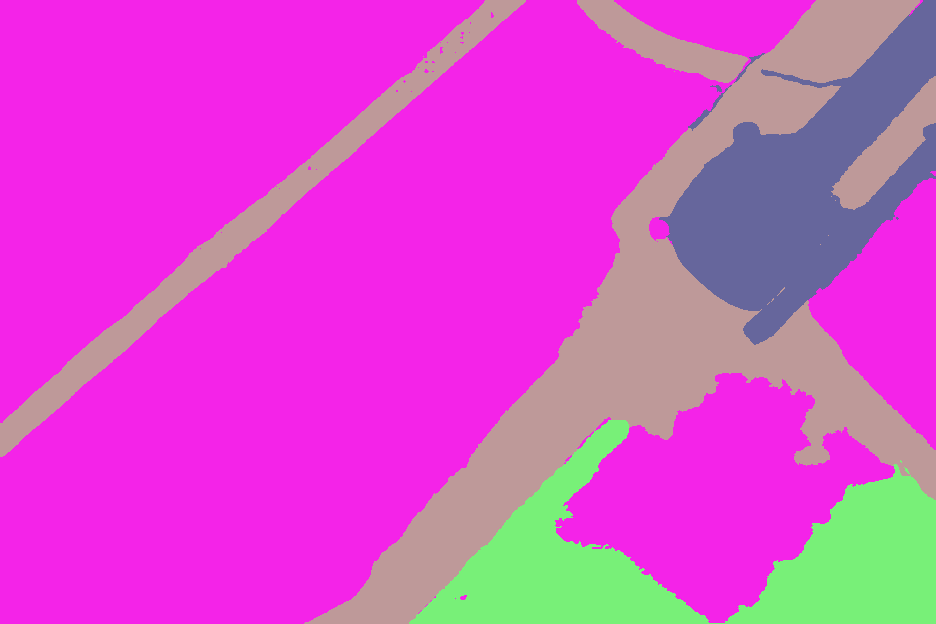}
    \end{subfigure}
    \begin{subfigure}{0.18\linewidth}
        \includegraphics[width=\linewidth]{images/SkyScapes/gt_1.png}
    \end{subfigure}

    \vspace{2mm}

    % Row 3
    \begin{subfigure}{0.18\linewidth}
        \includegraphics[width=\linewidth]{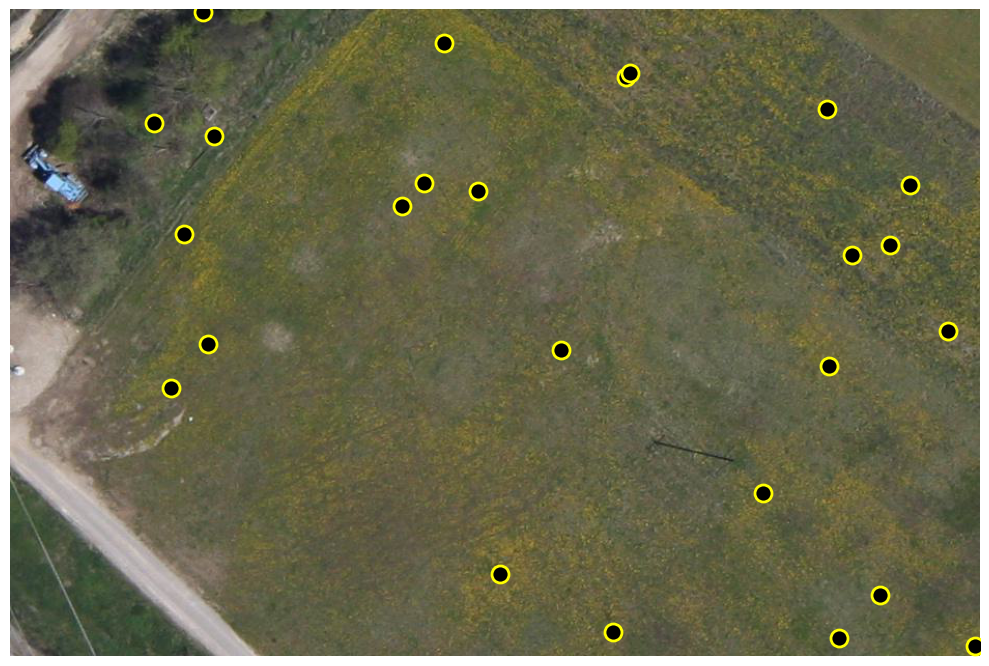}
    \end{subfigure}
    \begin{subfigure}{0.18\linewidth}
        \includegraphics[width=\linewidth]{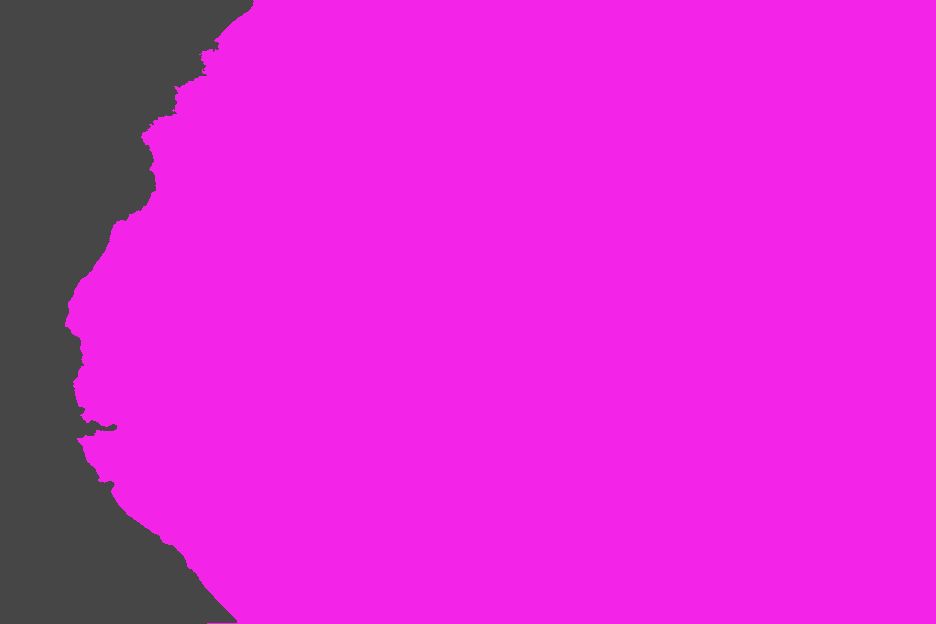}
    \end{subfigure}
    \begin{subfigure}{0.18\linewidth}
        \includegraphics[width=\linewidth]{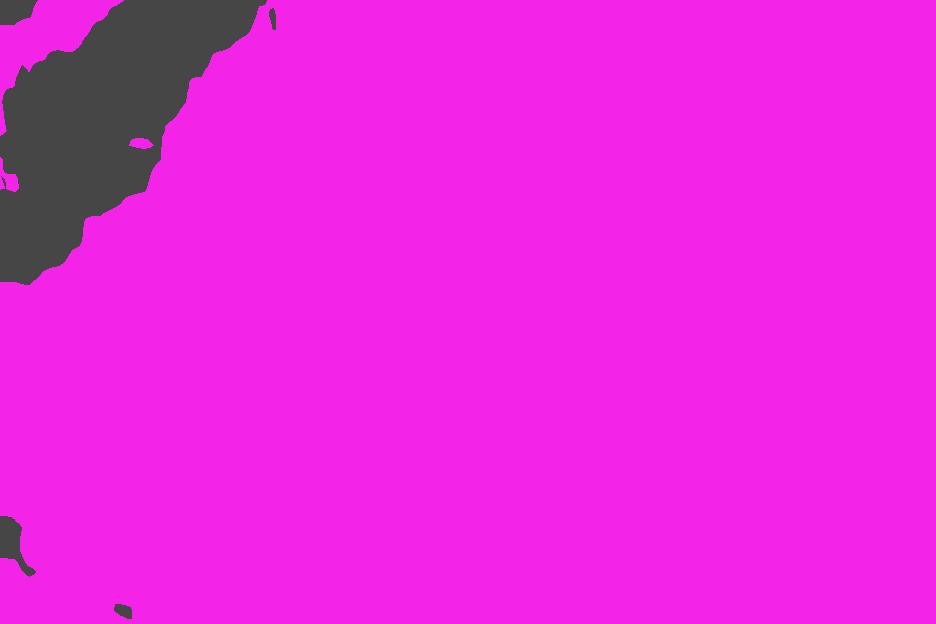}
    \end{subfigure}
    \begin{subfigure}{0.18\linewidth}
        \includegraphics[width=\linewidth]{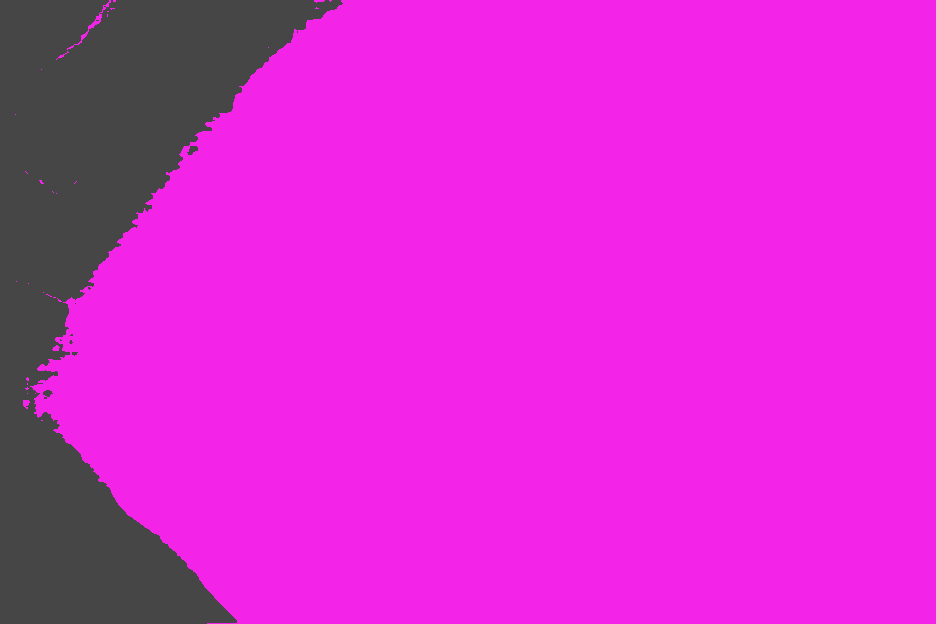}
    \end{subfigure}
    \begin{subfigure}{0.18\linewidth}
        \includegraphics[width=\linewidth]{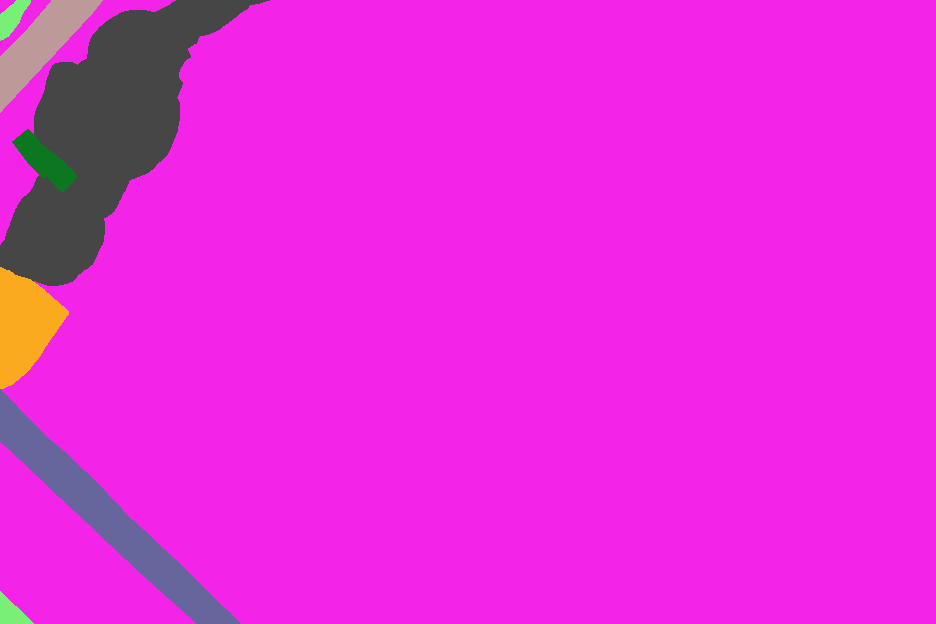}
    \end{subfigure}

    % Row 4
    \begin{subfigure}{0.18\linewidth}
        \includegraphics[width=\linewidth]{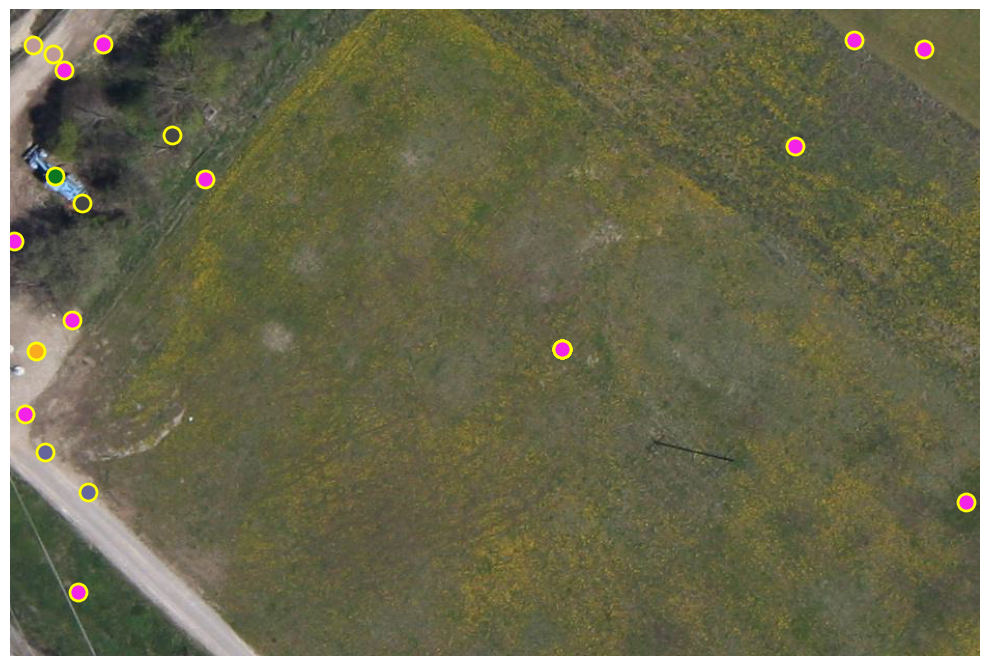}
    \end{subfigure}
    \begin{subfigure}{0.18\linewidth}
        \includegraphics[width=\linewidth]{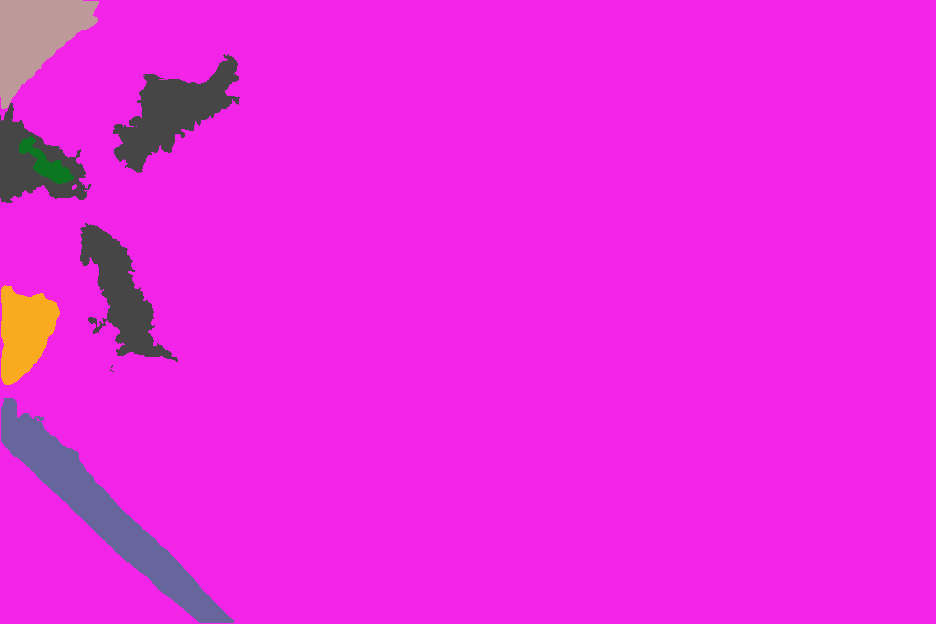}
    \end{subfigure}
    \begin{subfigure}{0.18\linewidth}
        \includegraphics[width=\linewidth]{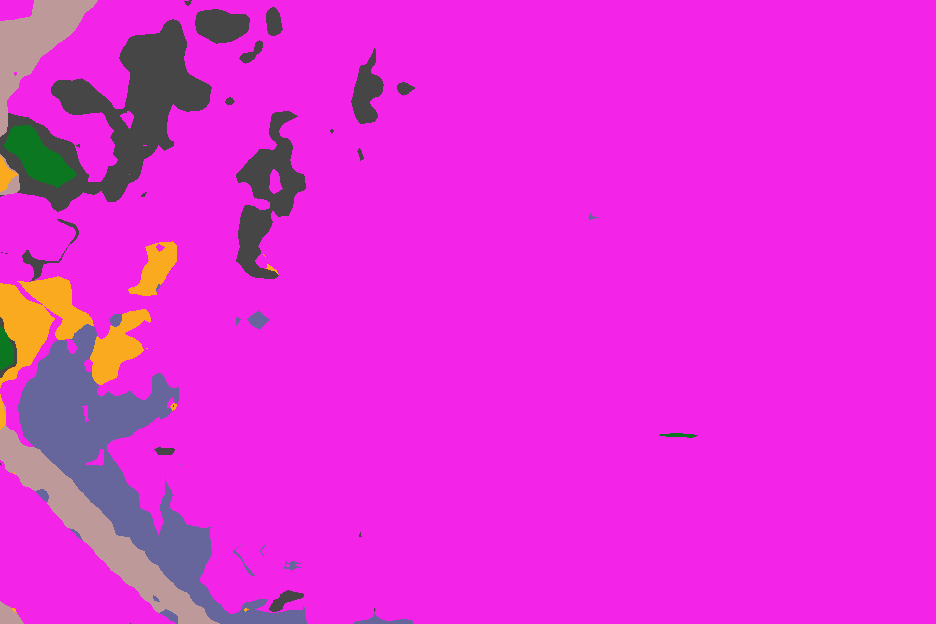}
    \end{subfigure}
    \begin{subfigure}{0.18\linewidth}
        \includegraphics[width=\linewidth]{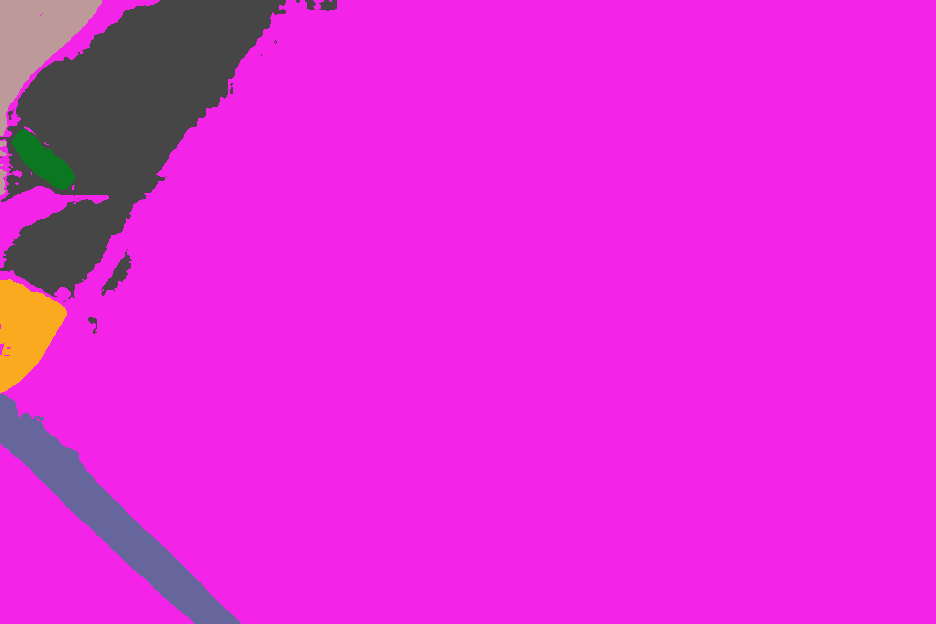}
    \end{subfigure}
    \begin{subfigure}{0.18\linewidth}
        \includegraphics[width=\linewidth]{images/SkyScapes/gt_2.png}
    \end{subfigure}

    \vspace{2mm}

    % Row 5
    \begin{subfigure}{0.18\linewidth}
        \includegraphics[width=\linewidth]{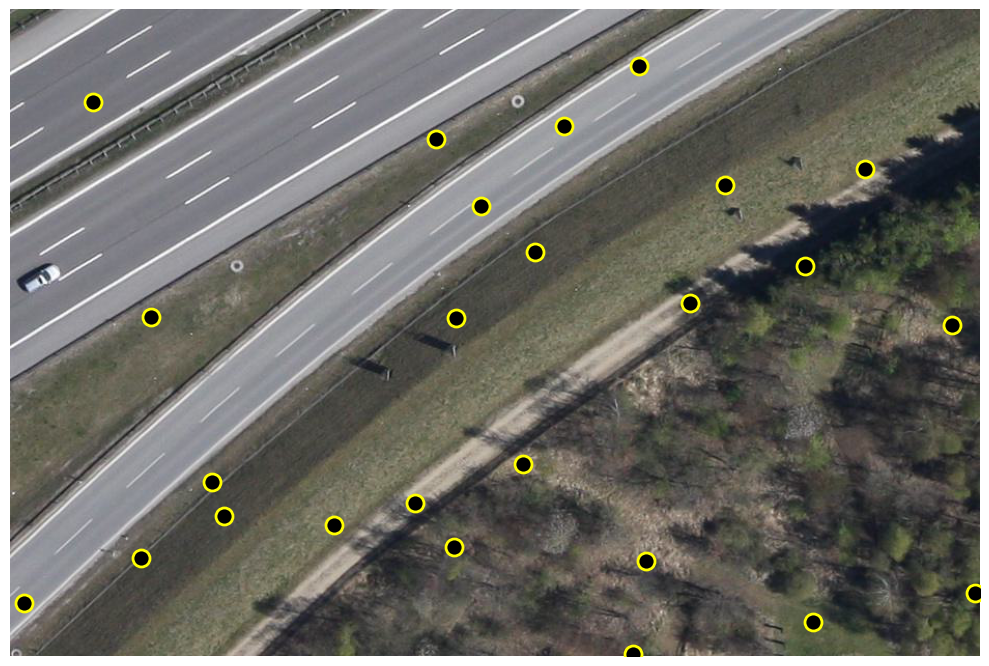}
    \end{subfigure}
    \begin{subfigure}{0.18\linewidth}
        \includegraphics[width=\linewidth]{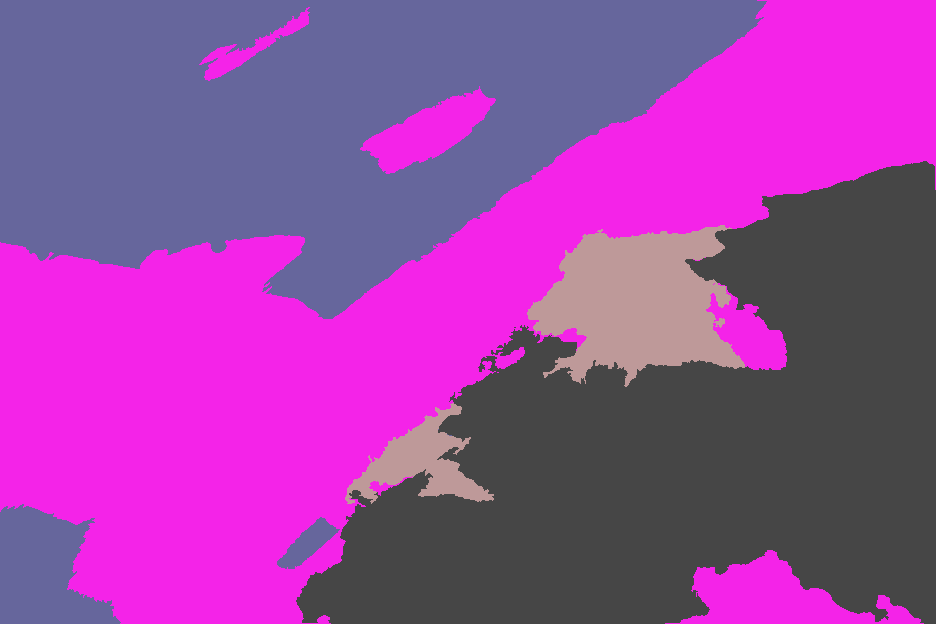}
    \end{subfigure}
    \begin{subfigure}{0.18\linewidth}
        \includegraphics[width=\linewidth]{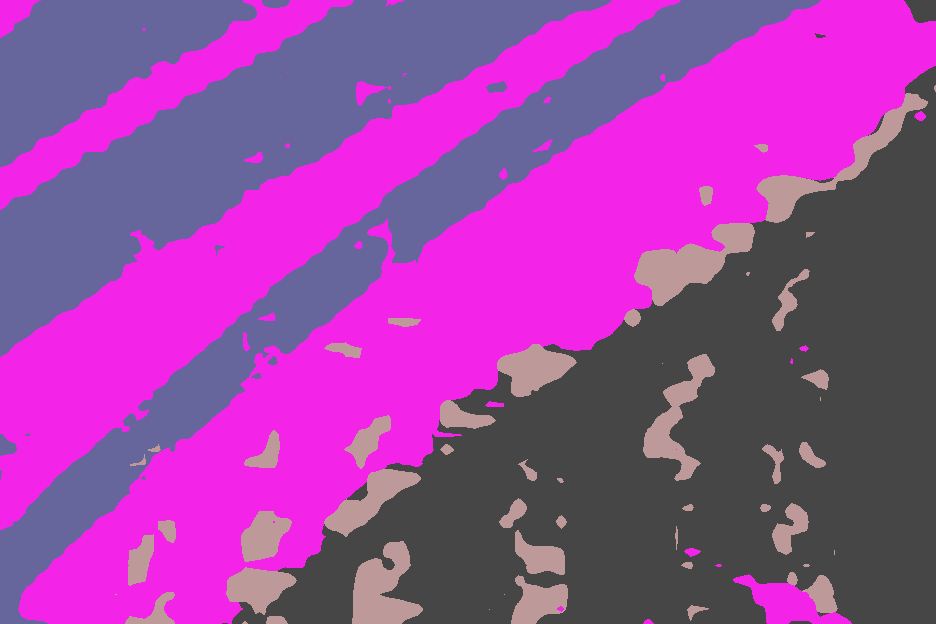}
    \end{subfigure}
    \begin{subfigure}{0.18\linewidth}
        \includegraphics[width=\linewidth]{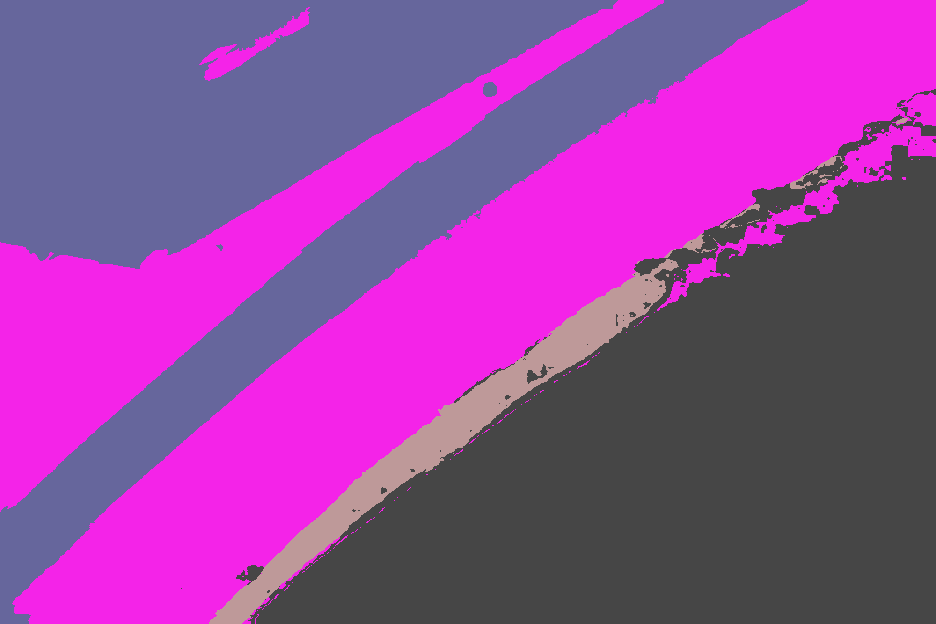}
    \end{subfigure}
    \begin{subfigure}{0.18\linewidth}
        \includegraphics[width=\linewidth]{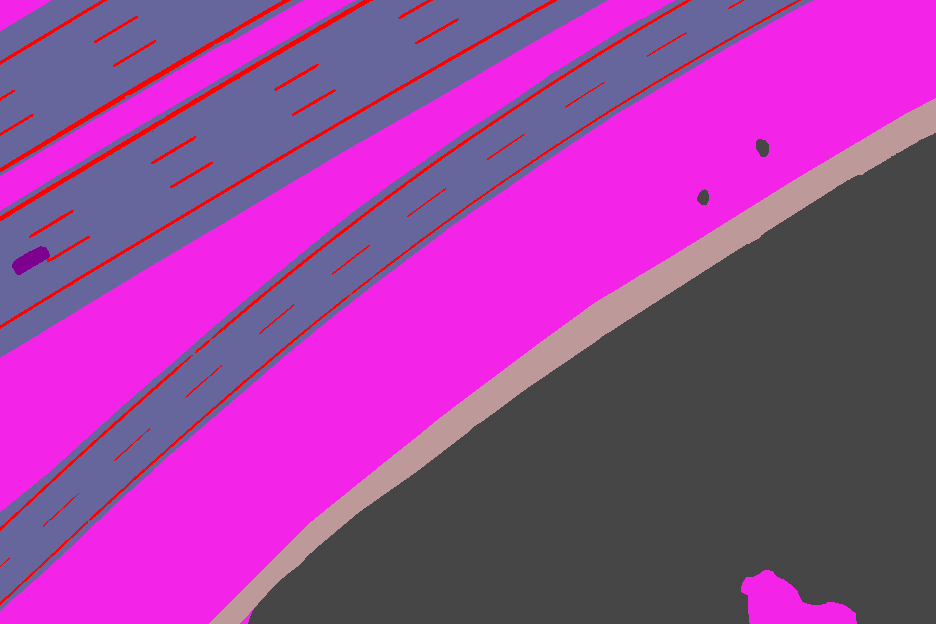}
    \end{subfigure}

    % Row 6
    \begin{subfigure}{0.18\linewidth}
        \includegraphics[width=\linewidth]{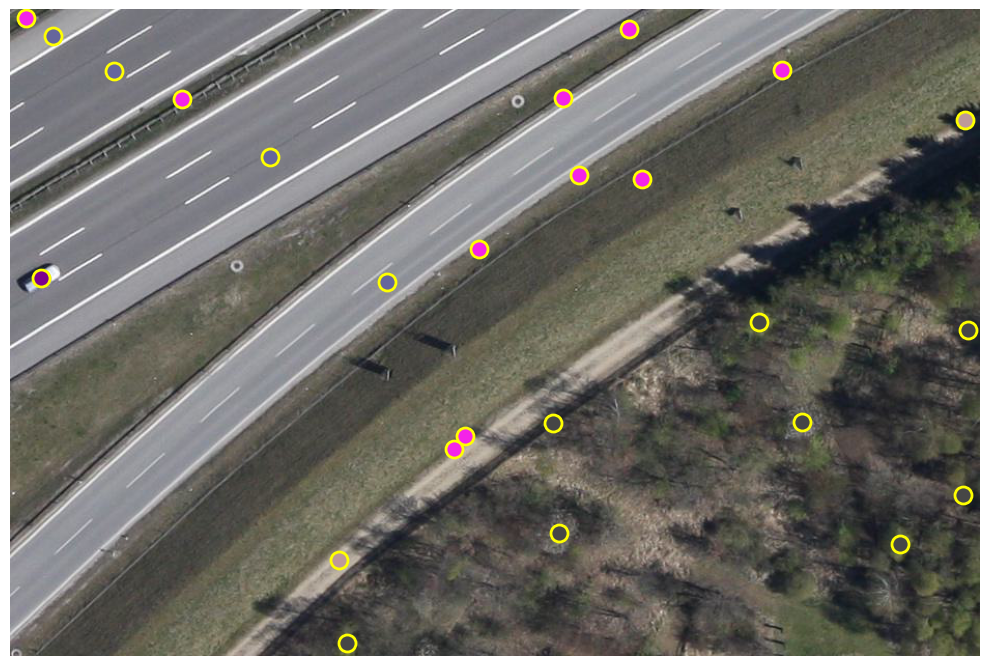}
    \end{subfigure}
    \begin{subfigure}{0.18\linewidth}
        \includegraphics[width=\linewidth]{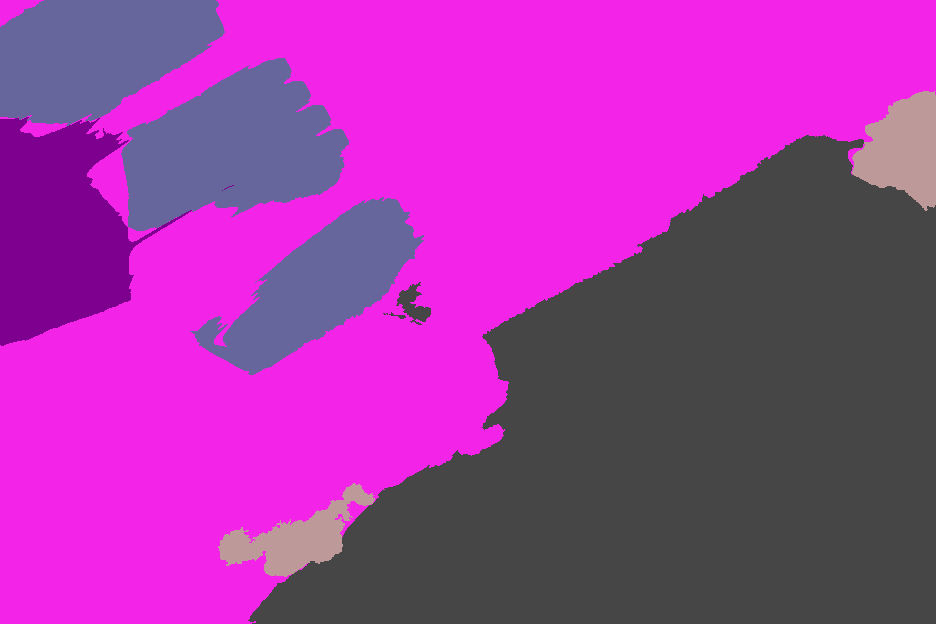}
    \end{subfigure}
    \begin{subfigure}{0.18\linewidth}
        \includegraphics[width=\linewidth]{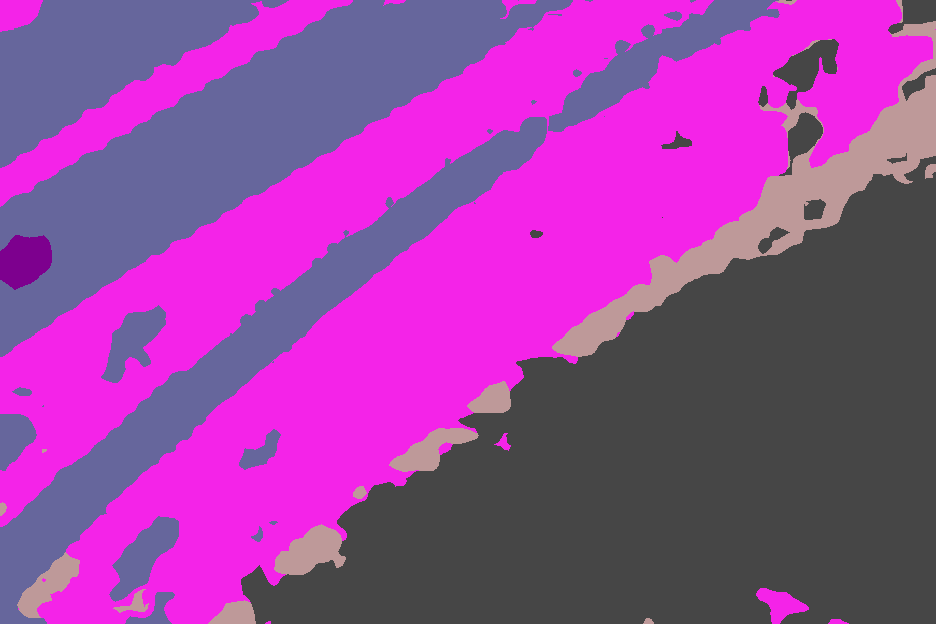}
    \end{subfigure}
    \begin{subfigure}{0.18\linewidth}
        \includegraphics[width=\linewidth]{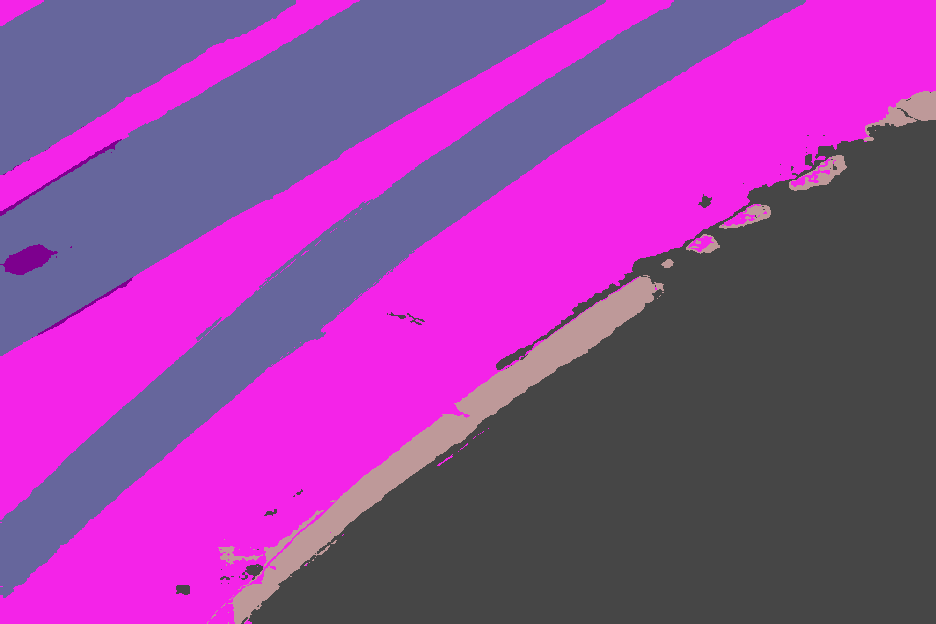}
    \end{subfigure}
    \begin{subfigure}{0.18\linewidth}
        \includegraphics[width=\linewidth]{images/SkyScapes/gt_4.png}
    \end{subfigure}

    \vspace{2mm}

    % Row 5
    \begin{subfigure}{0.18\linewidth}
        \includegraphics[width=\linewidth]{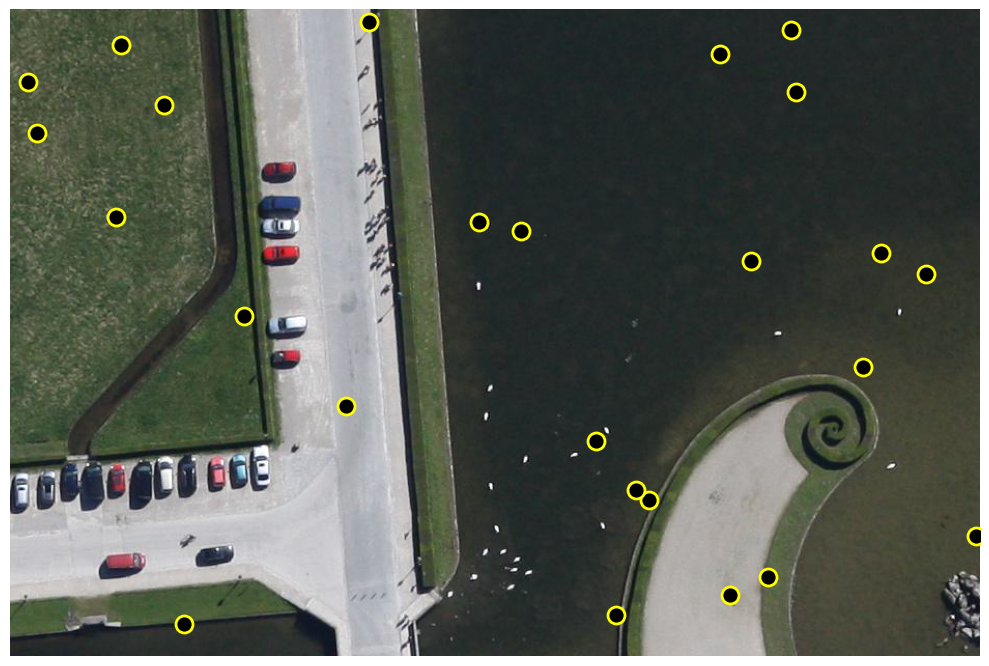}
    \end{subfigure}
    \begin{subfigure}{0.18\linewidth}
        \includegraphics[width=\linewidth]{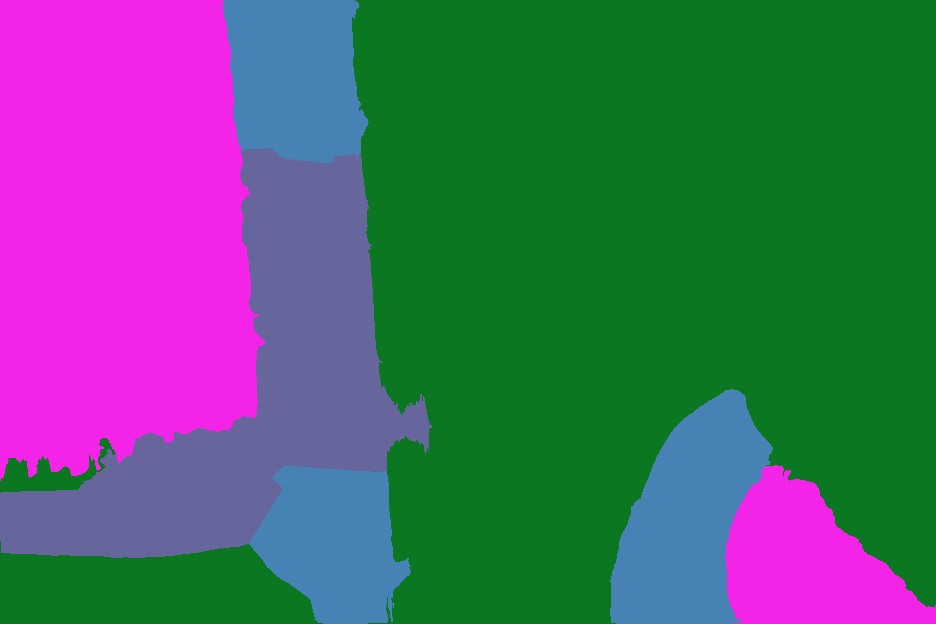}
    \end{subfigure}
    \begin{subfigure}{0.18\linewidth}
        \includegraphics[width=\linewidth]{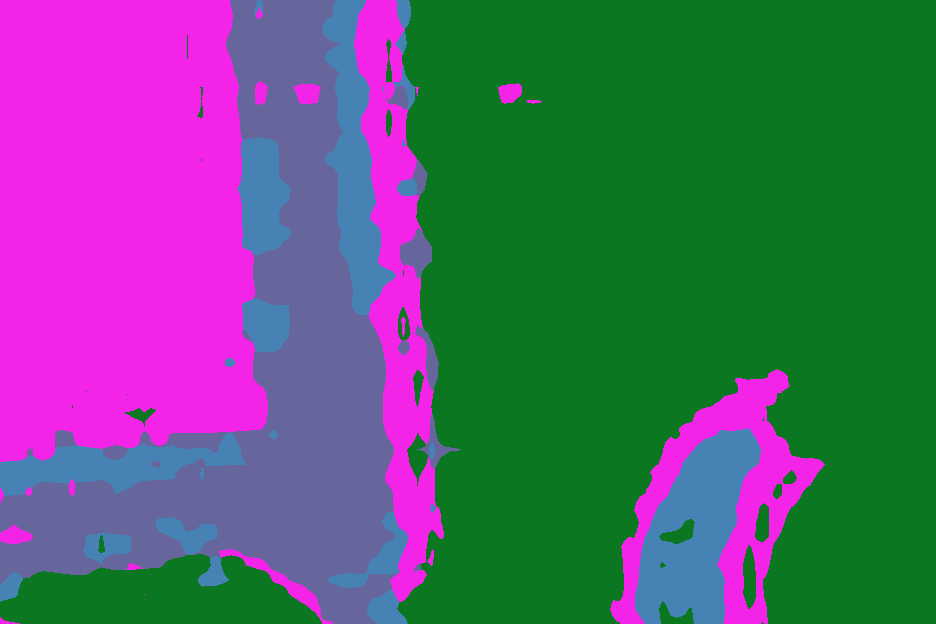}
    \end{subfigure}
    \begin{subfigure}{0.18\linewidth}
        \includegraphics[width=\linewidth]{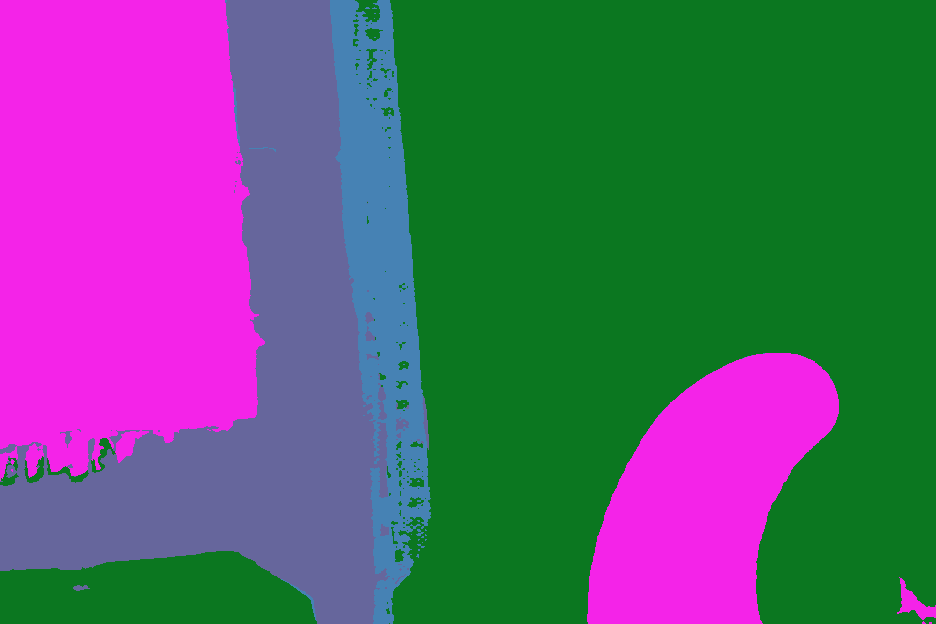}
    \end{subfigure}
    \begin{subfigure}{0.18\linewidth}
        \includegraphics[width=\linewidth]{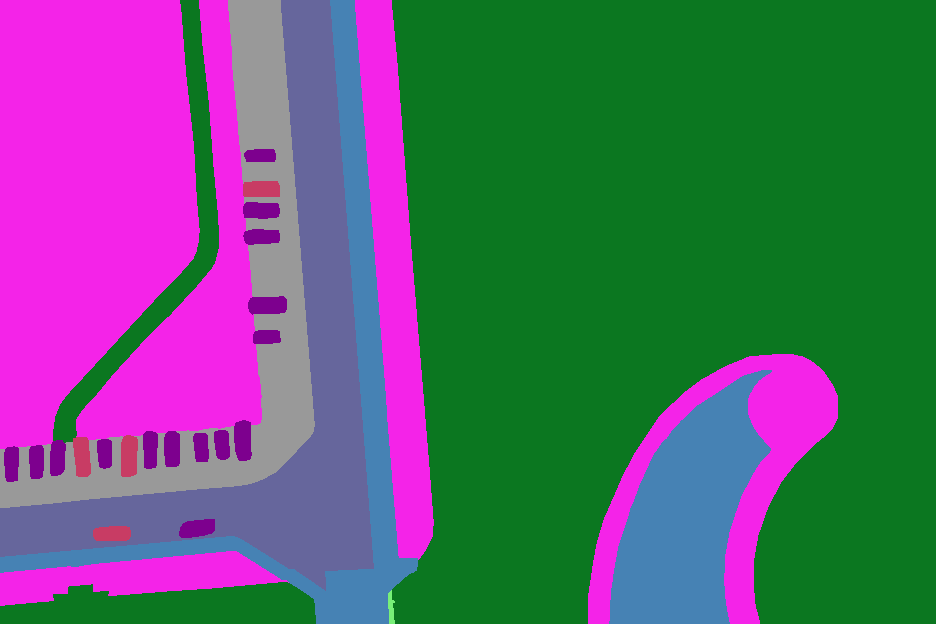}
    \end{subfigure}

    % Row 6
    \begin{subfigure}{0.18\linewidth}
        \includegraphics[width=\linewidth]{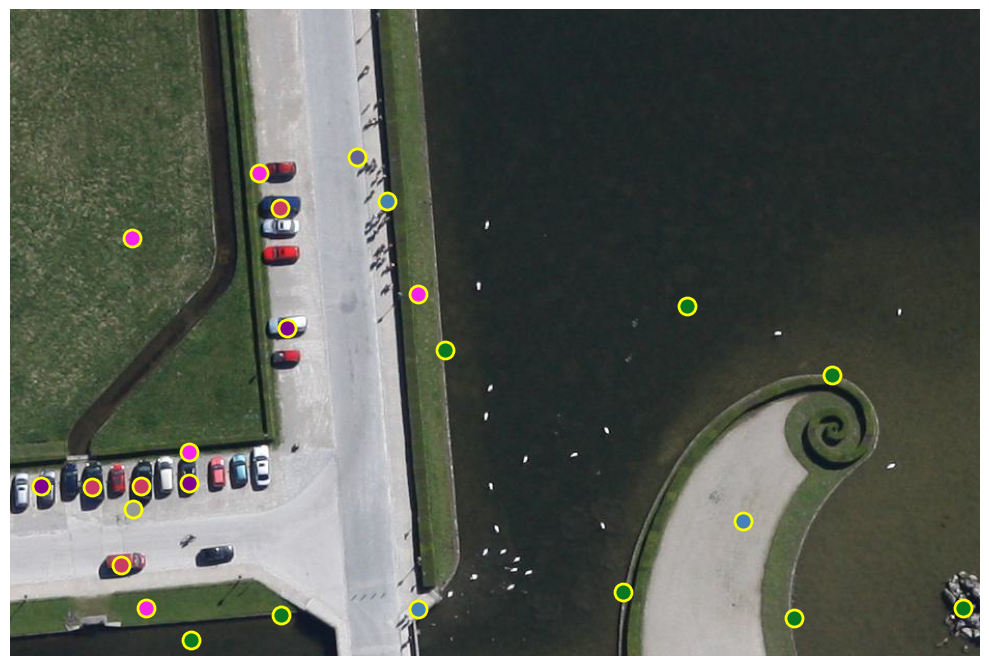}
        \caption*{\footnotesize (a) Img+PL}
    \end{subfigure}
    \begin{subfigure}{0.18\linewidth}
        \includegraphics[width=\linewidth]{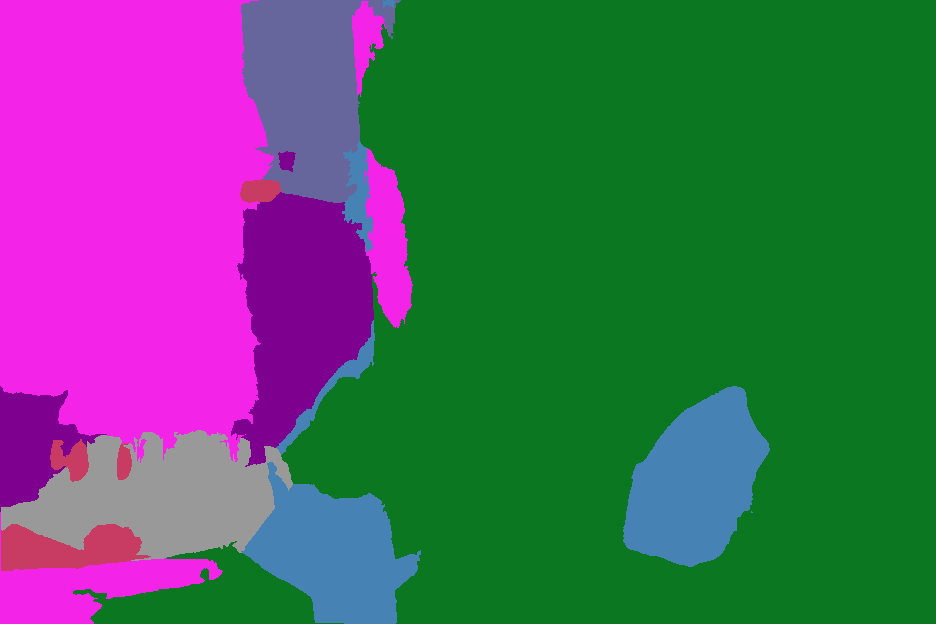}
        \caption*{\footnotesize (b) PLAS}
    \end{subfigure}
    \begin{subfigure}{0.18\linewidth}
        \includegraphics[width=\linewidth]{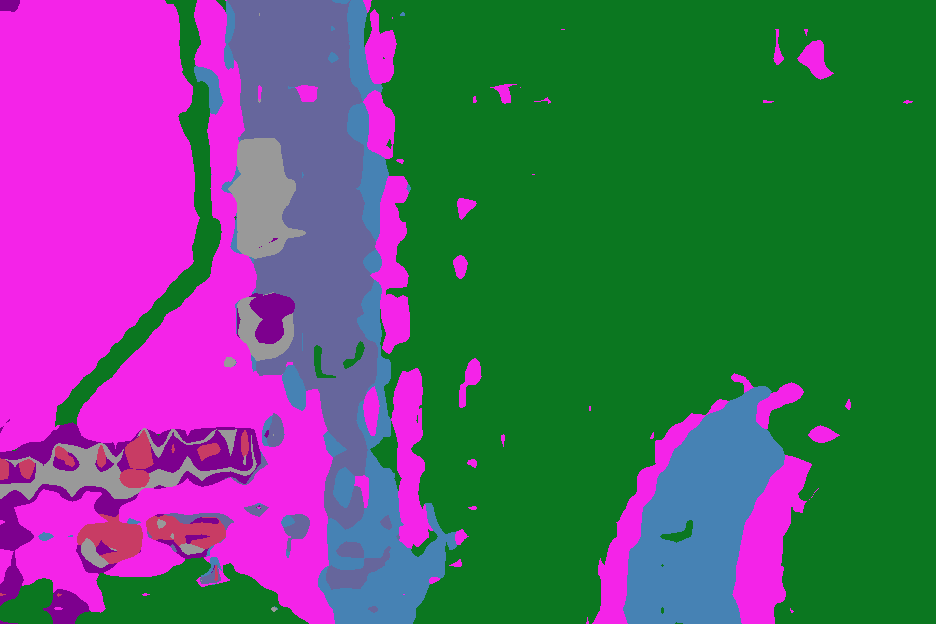}
        \caption*{\footnotesize (c) D+NN}
    \end{subfigure}
    \begin{subfigure}{0.18\linewidth}
        \includegraphics[width=\linewidth]{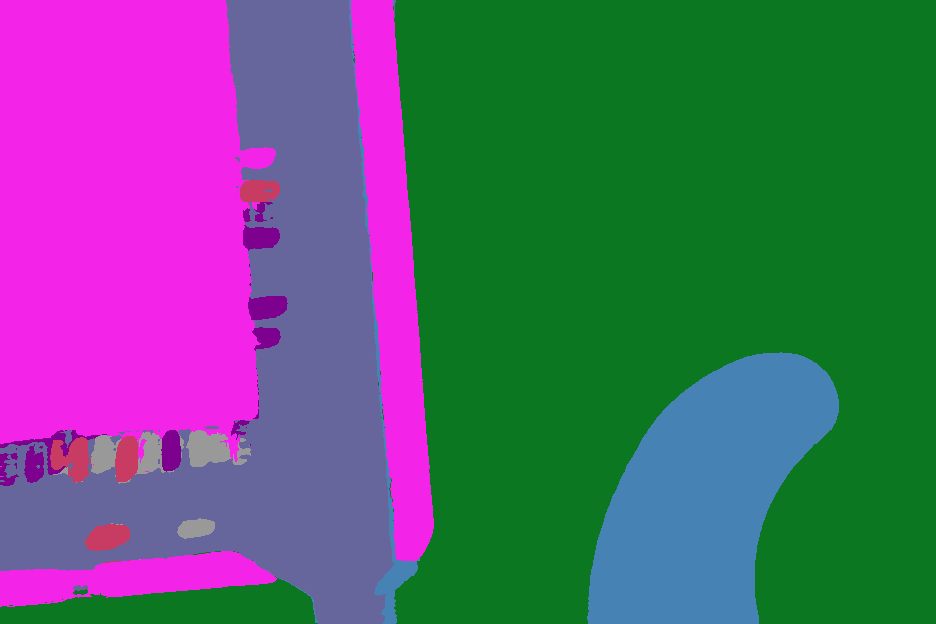}
        \caption*{\footnotesize (d) \textbf{Ours}}
    \end{subfigure}
    \begin{subfigure}{0.18\linewidth}
        \includegraphics[width=\linewidth]{images/SkyScapes/gt_6.png}
        \caption*{\footnotesize (e) GT}
    \end{subfigure}

    \vspace{2mm}
    
    \caption{Additioinal qualitative results. Label augmentation using PLAS, D+NN and SSeg on SkyScapes. Each pair of rows shows the final segmentation of the same image, with random sampling on top and our DynamicPoints sampling at the bottom. In each block, from left to right: (a) Input image with point-labels, (b) PLAS, (c) D+NN, (d) SSeg (Ours), (e) Ground Truth. Our DynamicPoints sampling leads to a better point placement to support effective label augmentation.
    %\fcolorbox{black}{background}{\rule{0pt}{2pt}\rule{3pt}{0pt}} Background color.
    }
\label{fig:additional_skyscapes}
\end{figure*}

\end{document}